\documentclass[10pt,twocolumn,letterpaper]{article}
\usepackage[margin=0.72in,columnsep=0.25in]{geometry}
\usepackage[T1]{fontenc}
\usepackage[utf8]{inputenc}
\usepackage{mathptmx,microtype}
\usepackage{amsmath,amssymb,booktabs,tabularx,array,multirow}
\usepackage{graphicx,xcolor,enumitem,pifont}
\usepackage{tikz}
\usepackage{placeins,longtable}
\usetikzlibrary{arrows.meta,positioning,fit,calc}
\usepackage[font=small,labelfont=bf]{caption}
\usepackage[numbers,sort&compress]{natbib}
\usepackage{url}
\usepackage[colorlinks=true,linkcolor=blue!45!black,citecolor=blue!45!black,urlcolor=blue!45!black]{hyperref}
\setlist[itemize]{leftmargin=*,itemsep=2pt,topsep=3pt}
\setlist[enumerate]{leftmargin=*,itemsep=2pt,topsep=3pt}
\newcommand{\fridge}{\textsc{Fridge}}
\newcommand{\cabinet}{\textsc{Cabinet}}
\newcommand{\sink}{\textsc{Sink}}

\newcommand{\code}[1]{\texttt{#1}}
\newcommand{\finding}[1]{\paragraph{#1}}
\newcommand{\insight}[2]{\par\smallskip\noindent{\setlength{\fboxsep}{5pt}\fcolorbox{blue!40!black}{blue!4}{\parbox{\dimexpr\columnwidth-2\fboxsep-2\fboxrule\relax}{\small\textbf{Takeaway #1.}\ #2}}}\par\smallskip}
\hypersetup{pdftitle={What Stops Recursive Self-Improvement in Robotics? Lessons from 123 Rounds of Agentic Skill Discovery},pdfauthor={Wang Jiaming}}
\begin{document}
\twocolumn[{
\begin{center}
{\LARGE\bfseries What Stops Recursive Self-Improvement in Robotics?\\[3pt]
Lessons from 123 Rounds of Agentic Skill Discovery\par}
\vspace{9pt}
{\large Wang Jiaming}\par
\vspace{2pt}
{\small National University of Singapore}\par
\vspace{3pt}
{\small Technical report}\par
\end{center}
\vspace{3pt}
\begin{center}\begin{minipage}{0.94\textwidth}
\small\textbf{Abstract.}
Can a robot improve itself the way coding agents now improve software?
We built an agentic system to find out. It watches a robot fail, works out which capability is missing, writes new skills or finds and installs external models, tests every change in simulation, and repeats, with no human writing robot code.
We ran it for 123 improvement rounds on household manipulation tasks. This report describes what we learned.
The good news is that the agent can discover capabilities on its own: noticing that its targets were out of view, it asked for an active-viewing model, debugged it, and deployed a working search skill.
The bad news is that its improvements did not add up. Changes kept passing their tests, yet the target task, putting condiments on the top shelf of a fridge, never succeeded.
We found that the agent was rarely the bottleneck. Three things around it were.
First, \emph{chained perception modules do not understand relations}. Segmenters such as SAM~3 find shelves but not ``the top shelf'', so the agent filled the gap with ever more geometric rules that never converged, when what it needed was a different kind of model.
Second, \emph{skill chains lock learning onto the first step}. Long tasks mostly fail early, so evidence and fixes pile up there, and later skills are rarely reached, tested, or improved.
Third, \emph{what the agent learns is decided by the harness}. The agent optimized exactly what the evaluator measured, including where it was wrong, and weak tests and misleading memory turned activity into a standstill.
We distill these lessons into concrete recommendations for building robot systems that improve themselves, each paired with an experiment that could prove it wrong.
\end{minipage}\end{center}
\vspace{9pt}
}]
\section{Introduction}
We want robots that get better on their own.
Today, when a robot fails at a household task, a person watches the video, works out what went wrong, and writes new code or collects new data. Coding agents suggest a different route. They can already read logs, form hypotheses, write and test code, and keep going for hours. Such agents already improve systems in loops: Voyager builds a growing library of Minecraft skills~\citep{wang2023voyager}, Eureka evolves reward functions for robot learning~\citep{ma2023eureka}, and the Darwin G\"odel Machine rewrites its own agent code~\citep{zhang2025darwin}. The idea has now reached robots: RHO optimizes robot-policy code repositories~\citep{elmaaroufi2026rho}, ASPIRE grows a library of robot skills from diagnosed failures~\citep{lu2026aspire}, and ENPIRE lets coding agents improve real-world robot policies~\citep{xiao2026enpire}.
So the idea is appealing. Let an agent watch the robot fail, figure out which capability is missing, write it or find a model that provides it, test it, and repeat until the task works, with no human writing robot code.

We built such a system and tested this idea (\S\ref{sec:system}).
Coding agents write the task programs, improve a library of robot skills, and research and install external perception and planning models. Every change must pass physical tests in simulation before the robot uses it. Humans were allowed to fix only the testing harness, never a skill.
We pointed it at a household task in RoboCasa~\citep{nasiriany2024robocasa}, placing condiments on the top shelf of a fridge, and let it run for 123 improvement rounds, several weeks of continuous operation.

We went in with three questions:
\begin{enumerate}
\item Can a coding agent \emph{discover} a missing robot capability from failures alone?
\item Do the improvements it makes \emph{add up}, so that each round starts from a better robot?
\item If not, \emph{what stops them}, and is it something a stronger agent would fix?
\end{enumerate}

The answer to the first question is yes, sometimes. The agent noticed that its targets were out of view, asked for an active-viewing model, debugged it, and deployed a search skill with no human robot code (\S\ref{sec:results}).
The answer to the second question is no. The system kept producing improvements that passed their tests, yet the fridge task never succeeded, and in the last twenty rounds the robot failed at the same step every time.
Most of this report is about the third question. We found that the agent was rarely the bottleneck. What stopped progress was what surrounded it. We organize what we learned around one observation (\S\ref{sec:framework}): an improvement helps only if the failure is \emph{described correctly} to the agent, the fix is \emph{within what the agent can change}, and the test can \emph{tell} whether it helped. In software all three come almost for free; in robotics each one broke, in a way that taught us something:

\begin{itemize}
\item \textbf{Lesson 1: chained perception modules do not understand relations} (\S\ref{sec:perception}). Household instructions say things like ``the top shelf''. SAM~3 can find shelves, but not \emph{which} shelf. Chaining it with other segmenters and depth does not help, because the relation lives between the modules. The agent, which can only write code, filled that gap with ever more geometric rules that never converged. What was needed was a different kind of model, a vision-language model, and switching model class was not an action the agent could take.
\item \textbf{Lesson 2: skill chains lock learning onto the first step} (\S\ref{sec:composition}). A fifteen-step task mostly fails early, so almost all evidence and almost all fixes concentrate on the first step. Later skills are rarely reached, rarely tested, and their improvements never reach the task.
\item \textbf{Lesson 3: what the agent learns is decided by the harness} (\S\ref{sec:learning}). The agent optimized exactly what the evaluator measured, including where the evaluator was wrong. Weak tests, an unchanging stream of evidence and a memory that recorded untried ideas as failures turned activity into a standstill. Keeping this harness honest took about two human repairs for every change the agent got accepted.
\end{itemize}

We close with what we would do differently (\S\ref{sec:design}) and how to test each suggestion. This is a report of one long experiment, not a controlled study (\S\ref{sec:limitations}). We think it is useful precisely because it failed in instructive ways that anyone building self-improving robots is likely to meet.

\begin{figure*}[t]
\centering
\includegraphics[width=\textwidth]{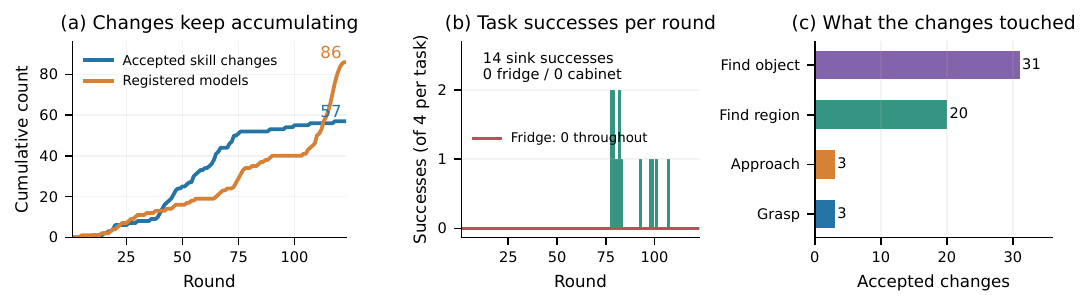}
\caption{\textbf{A busy loop that did not solve its task.}
(a) Accepted skill changes and newly registered models keep accumulating over 123 rounds.
(b) Task successes per round: none on the target \fridge\ task; the few on \sink\ were mostly accidental.
(c) Almost all accepted changes are to perception, the first step of every task (Lesson~2).}
\label{fig:campaign}
\end{figure*}

\section{System overview}
\label{sec:system}

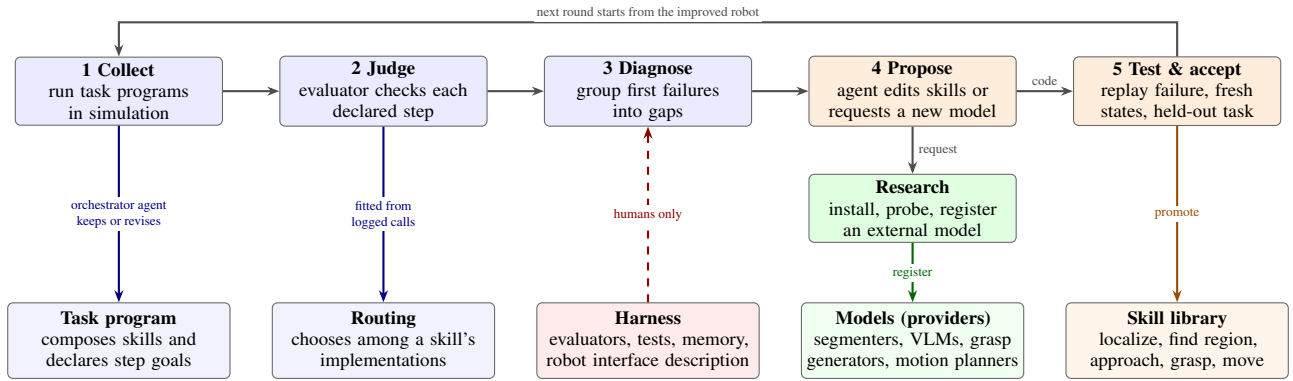
\begin{figure*}[t]
\centering
\begin{tikzpicture}[font=\scriptsize,
  step/.style={draw=black!60,rounded corners=2pt,fill=#1,align=center,minimum height=9mm,text width=26mm,inner sep=2pt},
  comp/.style={draw=black!60,rounded corners=2pt,fill=#1,align=center,minimum height=10mm,text width=28mm,inner sep=2pt},
  ar/.style={-{Stealth[length=1.6mm]},black!70,thick},
  up/.style={-{Stealth[length=1.6mm]},#1,thick},
  lab/.style={font=\tiny,fill=white,inner sep=1pt,align=center}]
\node[step=blue!8]   (collect) at (1.4,0)  {\textbf{1 Collect}\\run task programs\\in simulation};
\node[step=blue!8]   (judge)   at (4.9,0)  {\textbf{2 Judge}\\evaluator checks each\\declared step};
\node[step=blue!8]   (mine)    at (8.4,0)  {\textbf{3 Diagnose}\\group first failures\\into gaps};
\node[step=orange!14](propose) at (11.9,0) {\textbf{4 Propose}\\agent edits skills or\\requests a new model};
\node[step=orange!14](gate)    at (15.4,0) {\textbf{5 Test \& accept}\\replay failure, fresh\\states, held-out task};
\node[step=green!12] (research) at (11.9,-1.55) {\textbf{Research}\\install, probe, register\\an external model};
\draw[ar] (collect)--(judge); \draw[ar] (judge)--(mine); \draw[ar] (mine)--(propose);
\draw[ar] (propose)--node[lab,above=1pt]{code}(gate);
\draw[ar] (propose)--node[lab,right=1pt]{request}(research);
\draw[ar] (gate.north) -- ++(0,0.45) -| node[lab,pos=.25,above]{next round starts from the improved robot} (collect.north);
\node[comp=blue!5]   (prog)  at (1.4,-3.3)  {\textbf{Task program}\\composes skills and\\declares step goals};
\node[comp=blue!5]   (route) at (4.9,-3.3)  {\textbf{Routing}\\chooses among a skill's\\implementations};
\node[comp=red!8]    (harn)  at (8.4,-3.3)  {\textbf{Harness}\\evaluators, tests, memory,\\robot interface description};
\node[comp=green!6]  (prov)  at (11.9,-3.3) {\textbf{Models (providers)}\\segmenters, VLMs, grasp\\generators, motion planners};
\node[comp=orange!8] (skill) at (15.4,-3.3) {\textbf{Skill library}\\localize, find region,\\approach, grasp, move};
\draw[up=blue!50!black] (collect.south) -- node[lab]{orchestrator agent\\keeps or revises} (prog.north);
\draw[up=blue!50!black] (judge.south) -- node[lab]{fitted from\\logged calls} (route.north);
\draw[up=red!55!black,dashed] (harn.north) -- node[lab]{humans only} (mine.south);
\draw[up=green!40!black] (research.south) -- node[lab]{register} (prov.north);
\draw[up=orange!60!black] (gate.south) -- node[lab]{promote} (skill.north);
\end{tikzpicture}
\caption{\textbf{How the system improves itself.} Each improvement round (top row) runs the robot, judges every declared step, groups failures, and lets a coding agent (a GPT model in the Codex CLI, shown camera keyframes of each failure) propose a fix, either as code that must pass physical tests or as a request for a new external model. Five parts of the robot can change (bottom row); arrows show who changes each one. Four are changed by agents or by learning from logs. The harness that judges and remembers is changed only by humans, which turns out to matter (\S\ref{sec:learning}).}
\label{fig:system}
\end{figure*}

The idea is simple: treat the robot's software as something an agent can rewrite, and let failures tell it what to rewrite.
Figure~\ref{fig:system} shows the system. The robot is controlled by a task program that calls skills. Skills call external models. Every part except the evaluation harness can be changed by an agent or by learning, without a human writing robot code.

\finding{The robot's software.}
A \emph{task program} is Python code that composes skills and declares what each step should achieve, for example ``the top shelf inside the fridge is localized'', ``the object is grasped and lifted'', or ``the object is placed''.
A \emph{skill} (localize an object, find a region, approach, grasp, move the arm) has typed inputs and outputs and can have several implementations. \emph{Routing} rules choose among them.
Skills can call \emph{providers}: external models behind a contract, such as text-prompted segmentation (SAM~3~\citep{carion2025sam3}), referring-expression segmentation (Florence-2~\citep{xiao2024florence}), monocular depth, 6-DoF grasp generation or motion planning.
The robot perceives only through its head and wrist RGB-D cameras and its joint readings, as a real robot would.

\finding{How each part improves.}
\emph{Programs} are written by an orchestration agent from the task description. It keeps its best past program and revises it only for program faults, so that skill failures are fixed in skills rather than worked around.
\emph{Skills} are improved by a proposer agent. For each selected gap, it sees the recorded failures, the evaluator's diagnosis and its own earlier attempts, and writes a code change.
\emph{Providers} are added by a research agent. When the proposer decides that no code change can fix a failure, for example because a model is missing, it files a request. The research agent then finds a model, installs it, checks it on a probe scene and on recorded failing inputs, and registers it.
\emph{Routing} is refitted every round from logged skill calls.
The \emph{harness}, meaning the evaluators, tests, memory and the description of the robot's actuators, is the one part only humans may change. Our rule was that humans can repair the harness but never write a skill, program or model wrapper.

\finding{How a change is accepted.}
The evaluator judges each declared step with privileged simulator state and returns a short diagnosis, such as \code{wrong\_region}, \code{no\_grasp\_contact} or \code{tool\_path\_blocked}.
A proposed code change must pass a sequence of physical tests. First, it replays the exact recorded failure with the change swapped in; did it fix what it claims to fix? Second, it runs paired trials, old against new, on fresh scenes and on a held-out task the proposer never sees; does it help without hurting elsewhere? Later in the project we added tests on sealed recorded failures and on the other skills a change touches.
Accepted changes go live in the next round.

\finding{The agent models.}
All three agents (orchestrator, proposer and research agent) are OpenAI GPT models run through the Codex CLI, each session in its own sandboxed container (Table~\ref{tab:models}).
The proposer is multimodal. Along with the failure table, the evaluator's diagnosis and its earlier attempts, it receives RGB keyframes from the robot's cameras at the start and end of every failed step, attached as images. It can therefore \emph{look at} the failures it is asked to fix.
The model and its reasoning effort changed twice during the run, both times by our decision, not the agent's: from \code{gpt-5.6-sol} at maximum reasoning effort to \code{gpt-6-astra} at low effort (round~76), for speed, and then to medium effort (round~85).
The robot itself never calls these models. Its perception comes only from the providers its skills call.

\begin{table}[t]
\centering\small
\caption{\textbf{Agent models.} All agents used the same model and reasoning setting in a given round.}
\label{tab:models}
\setlength{\tabcolsep}{4pt}
\begin{tabularx}{\columnwidth}{@{}lX@{}}
\toprule
Rounds 1--75 & \code{gpt-5.6-sol}, maximum reasoning effort\\
Rounds 76--84 & \code{gpt-6-astra}, low reasoning effort\\
Rounds 85--123 & \code{gpt-6-astra}, medium reasoning effort\\
\midrule
Interface & Codex CLI, one session per agent task, in a read-only, capability-dropped container; from round~74 the run repository and held-out task are not visible\\
Orchestrator & Task description, skill catalog with observed usage, its best earlier programs; up to 25\,min per program\\
Proposer & Failure table, evaluator diagnoses, execution traces and diffs of earlier attempts, \emph{camera keyframes of each failed step}; up to 3 attempts per problem, 60\,min each\\
Research agent & The request and recorded failing inputs; finds, installs, wraps and tests an external model\\
\bottomrule
\end{tabularx}
\end{table}

\finding{Tasks.}
We used three RoboCasa~\citep{nasiriany2024robocasa} kitchen tasks with a PandaOmron mobile manipulator. The target task was \fridge: \emph{``Place the mustard and mayonnaise from the counter to the top shelf of the fridge. If the existing items in the fridge are on the top shelf, move them to other shelves.''} The condiments start out of view, and the task takes roughly fifteen steps.
Two simpler pick-and-place tasks, \cabinet\ (counter to cabinet) and \sink\ (counter to sink), were added later so that skills had to generalize across tasks. A fourth task, cabinet to counter, was held out for testing only.
A round took one to three hours, plus several hours when models were researched, and we ran 123 rounds.

\section{What happened over 123 rounds}
\label{sec:results}

\begin{figure*}[t]
\centering\includegraphics[width=\textwidth]{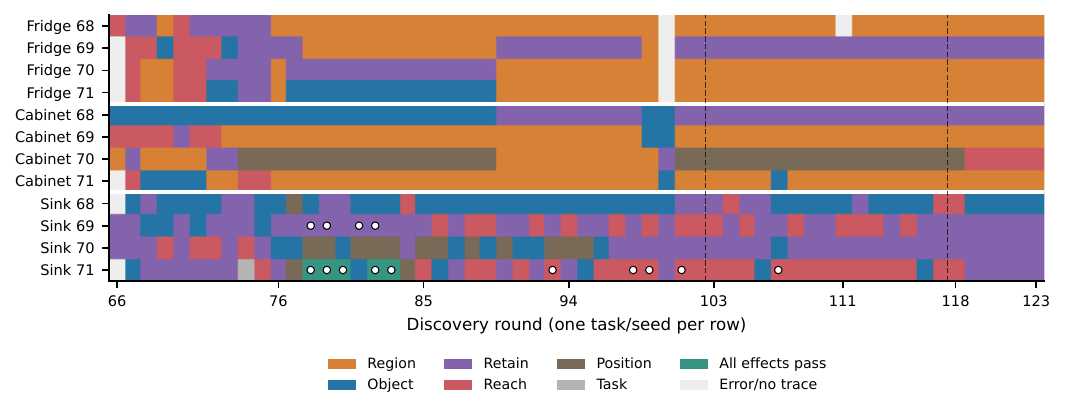}
\caption{\textbf{Where each rollout first failed, rounds 66--123.} Each row is one task and scene; color is the first step that failed. White circles mark task successes. Early in this window failures move around as skills change; in the last twenty rounds the fridge rows freeze at region finding (``the top shelf'') and grasping, and stay there despite continued library changes and new models.}
\label{fig:frontier}
\end{figure*}

Figure~\ref{fig:campaign} summarizes the whole run, and Figure~\ref{fig:frontier} shows, round by round, where each task first failed. The story has three phases.

\finding{Phase 1: discovery (rounds 1--36).}
Early on, the loop did what we hoped.
The condiments in \fridge\ start out of view, and nothing in any prompt said so. At first the localizer returned confident points on visible look-alikes, and the agent tried to re-rank them. Once we fixed the evaluator so that it reported ``the target is not visible from any camera'', the agent changed approach: it asked for a ``target-free, collision-aware active viewpoint selection'' model, found a bug in the planner it was given (the robot's own start cell was marked as occupied), requested a repair, and wrapped the result in a look-around skill. By round~16 the robot searched when it could not see its target, and out-of-view failures disappeared.
In the same phase it created a new skill family for finding regions, asked for a referring-expression segmenter, a grasp generator and motion planners, and, once the region evaluator was made truthful, produced a top-shelf fix that passed every test.

\finding{Phase 2: the manipulation front and the harness (rounds 37--102).}
With perception moving, failures shifted to reaching and grasping, and progress there exposed the harness rather than the agent.
The reach evaluator turned out to be a sphere, and the agent had learned to park exactly on its surface. The robot's torso lift was missing from its action description. The base controller behaved very differently from its documentation.
Each fix opened a new front. The first approach skill was accepted in round~62 and the first grasp skill in round~63; the \cabinet\ and \sink\ tasks were added so that skills had to work across tasks.
The first task successes, on \sink, came in round~78. They did not hold: in later rounds the same program mostly succeeded by accident, when a released object happened to fall into the sink. Meanwhile the top-shelf problem came back as programs and viewpoints drifted. By the end of this phase, the fridge and cabinet rollouts were identical from one round to the next.

\finding{Phase 3: more capability, same outcome (rounds 103--123).}
We repaired the loop's memory and tests: lost research requests were recovered, partial fixes were allowed to reach the fresh-scene tests, and small differences had to reproduce before they counted.
The research agent then registered new models every round, including the first vision-language grounders for the top-shelf problem. A grasp improvement in round~118 won reproduced comparisons on unseen scenes, including the held-out task.
None of it moved the target task. In every round of this phase, three of the four fridge scenes failed at finding the top shelf and the fourth at grasping. The new grounders were never called by the robot.

\finding{What worked.}
\begin{itemize}
\item \textbf{The agent can discover a missing capability} when the failure is described correctly. The active-search skill was invented, researched, debugged and deployed with no human robot code.
\item \textbf{The agent can acquire and adapt external models.} Eight agent-introduced models (grasp generation, motion and base planning, active viewing, relational localization) were used in deployment.
\item \textbf{Carrying partial designs forward works.} Designs converged once rejected attempts returned their execution traces and the best earlier design seeded the next attempt, instead of each attempt starting over.
\item \textbf{The agent is a good diagnostician.} When it declined to change a skill and explained why, it was often pointing at a real harness bug.
\end{itemize}

\finding{What did not work.}
\begin{itemize}
\item \textbf{The target task never succeeded.} Neither did the cabinet task; the sink successes were mostly accidental drops.
\item \textbf{Changes piled up at the front of the task.} Almost all accepted changes were to perception, the first step of every task (Fig.~\ref{fig:campaign}c).
\item \textbf{Local improvements did not add up.} Changes that passed rigorous tests, and new models that passed research, did not change what the robot achieved.
\item \textbf{Keeping the loop honest took constant human work}, about two harness repairs for every accepted skill change.
\end{itemize}
The rest of this report explains why, and what we would change.

\section{Why improvements did not add up}
\label{sec:framework}

\finding{A minimal model of the loop.}
Let $\Phi_t$ be the robot's library after round $t$ and let a task be a chain of $m$ declared steps.
Write $p_j(\Phi)$ for the probability that step $j$ succeeds given that steps $1,\dots,j-1$ did, and $\pi_j=\prod_{i<j}p_i$ for the probability that a rollout \emph{reaches} step $j$.
Task success is $S(\Phi)=\prod_{j=1}^{m}p_j$, which requires no independence assumption.
Each round, the loop observes failures, a proposer produces a change $\delta$, and a gate $G$ accepts it, so that $\Phi_{t+1}=\Phi_t\oplus\delta$ when $G(\delta)=1$.
Self-improvement \emph{compounds} when accepted changes raise $S$ in expectation, so that each round starts from a better robot than the last.

\finding{Three conditions.}
Whether an accepted change raises $S$ depends on three conditions. None of them concerns the proposer's coding ability.
\begin{enumerate}
\item \textbf{Representability.} The failure must appear in the evidence the proposer sees, attributed to its actual cause. If a grasp fails because the arm stopped short, and the evidence reads ``no grasp contact'', the proposer will improve the wrong thing, and do it well.
\item \textbf{Expressibility.} A fix must exist in the substrate the proposer is allowed to change. A proposer that can edit code but cannot change what a perception model understands can only approximate missing semantics with code.
\item \textbf{Resolvability.} The gate must be able to tell a better change from a worse one on the states where the change matters. A gate that sees too few relevant states, or noise larger than the effect, selects at random or by accident.
\end{enumerate}
In software self-improvement the three hold almost for free. A failing unit test states its cause, the fix lies in the code the agent edits, and the test is exact and cheap to re-run. In a robot loop each condition has to be engineered, and each broke in our experiment in a characteristic way.

\begin{table*}[t]
\centering\small
\caption{\textbf{The main problems of our run, through the three conditions.} \ding{51}~holds, \ding{55}~fails, $\circ$~holds only after harness repair. The one front where all three held is the one the agent solved on its own.}
\label{tab:fronts}
\renewcommand{\arraystretch}{1.15}
\begin{tabularx}{\textwidth}{@{}p{.2\textwidth}ccc X@{}}
\toprule
Front & Repr. & Expr. & Resol. & What happened\\
\midrule
Target outside every view (active search) & $\circ$ & \ding{51} & \ding{51} & Once the evaluator reported visibility, the agent requested an active-viewing model, repaired it, wrapped it in a look-around skill, and the failure disappeared from deployment (\S\ref{sec:results}).\\
``The top shelf'' (relational region) & $\circ$ & \ding{55} & \ding{55} & No module represents the relation; the agent encodes it as geometry fit to two recorded states. The skill grows without converging (\S\ref{sec:perception}).\\
Grasp retention & \ding{55} & \ding{51} & \ding{55} & Motion faults reported as grasp faults; fresh tests rarely reach the grasp. Real local wins never reach the task (\S\ref{sec:composition}).\\
Reachability & \ding{55} & \ding{51} & \ding{51} & A spherical evaluator and a misdescribed actuator; the agent optimizes the evaluator exactly (\S\ref{sec:learning}).\\
Requested models & \ding{55} & \ding{51} & \ding{55} & Requests silently dropped and remembered as failed; registered models never called (\S\ref{sec:learning}).\\
\bottomrule
\end{tabularx}
\end{table*}

\finding{Reading our run through the three conditions.}
Table~\ref{tab:fronts} places the main problems of the run on these axes. The one front where all three conditions held, active search, is the one the agent solved on its own (\S\ref{sec:results}). Its early failure was a representability failure: while the evaluator blamed the wrong cause, the agent kept improving the wrong thing. This tells us that the proposer was not the bottleneck. The next three sections are our three lessons about what is.

\section{Lesson 1: chained perception modules do not understand relations}
\label{sec:perception}

The fridge instruction reads: \emph{place the condiments on the top shelf; if items already occupy the top shelf, move them to other shelves.}
Every program the orchestrator wrote therefore asks perception for regions defined by relations: ``the top shelf inside the fridge'', ``the shelf immediately below the top shelf'', ``a food item on the top shelf''.
This is exactly the kind of query household instructions are made of, and the loop never learned to answer it. Region grounding became the most common first failure on both the fridge and the cabinet task. In the last twenty rounds, three of the four fridge seeds failed at it every round, typically by returning the shelf \emph{below} the top shelf, sometimes with high confidence. This section explains why a capable code-writing agent could not fix this, and why the way it tried is itself the problem.

\begin{figure}[t]
\centering
\begin{tikzpicture}[font=\scriptsize,
  st/.style={draw=black!55,rounded corners=2pt,fill=#1,align=center,minimum height=5.5mm,text width=15mm,inner sep=2pt},
  ar/.style={-{Stealth[length=1.4mm]},black!70}]
\node[font=\scriptsize\bfseries,anchor=west] at (-0.2,0.55) {(a) What the loop built: modules chained by glue};
\node[st=black!4] (q1) at (0.6,0) {``top shelf\\in the fridge''};
\node[st=blue!8,right=3mm of q1] (s1) {concept /\\referring seg.};
\node[st=blue!8,right=3mm of s1] (d1) {depth lift\\per mask};
\node[st=red!10,right=3mm of d1] (h1) {geometric glue\\(33 thresholds)};
\draw[ar] (q1)--(s1); \draw[ar] (s1)--(d1); \draw[ar] (d1)--(h1);
\node[below=0.6mm of s1,font=\tiny,text width=17mm,align=center] {\emph{what} a shelf is,\\not \emph{which}};
\node[below=0.6mm of h1,font=\tiny,text width=17mm,align=center,text=red!50!black] {relation decided\\by fitted rules};
\node[font=\scriptsize\bfseries,anchor=west] at (-0.2,-1.35) {(b) Reason first, then verify};
\node[st=black!4] (q2) at (0.6,-1.9) {``top shelf\\in the fridge''};
\node[st=green!12,right=3mm of q2] (v2) {VLM grounds\\points / boxes};
\node[st=blue!8,right=3mm of v2] (s2) {segment +\\depth lift};
\node[st=green!12,right=3mm of s2] (c2) {geometric\\veto; abstain\\$\to$ look};
\draw[ar] (q2)--(v2); \draw[ar] (v2)--(s2); \draw[ar] (s2)--(c2);
\node[below=0.6mm of v2,font=\tiny,text width=17mm,align=center] {relation resolved\\over the scene};
\node[below=0.6mm of c2,font=\tiny,text width=17mm,align=center,text=green!35!black] {geometry checks,\\does not decide};
\end{tikzpicture}
\caption{\textbf{Where the relation lives.} (a) With concept-level and referring-expression segmenters, no module represents ``top'' or ``below''; the relation is implemented in agent-written glue. (b) A vision-language model resolves the relation over the whole image; geometry and active viewing only verify or reject its answer.}
\label{fig:perception}
\end{figure}
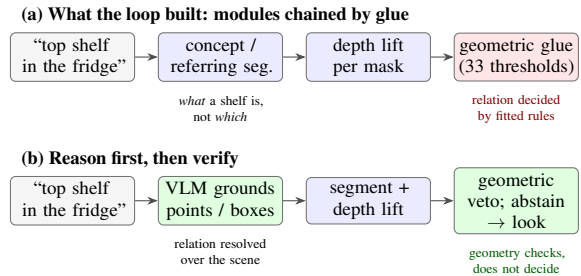

\finding{Modules do not compose into relations.}
The agent had two grounding models to build on: SAM~3~\citep{carion2025sam3}, prompted with text, and a Florence-2 referring-expression segmenter~\citep{xiao2024florence} that the agent itself requested in the first round. Both are strong models. Neither represents the part of the query that matters.
SAM~3 is built for \emph{concept} prompts, i.e.\ short noun phrases. It finds shelves, all of them, and has no notion of which one is ``top'' or ``first''.
Florence-2 accepts the full expression, but for relational queries it returned diffuse masks spanning several shelves.
Chaining the two with depth does not help, because depth adds metric position without adding meaning.
The relational part of the query (ordinal position, ``below'', ``inside'', ``occupied by'') is therefore represented by \emph{none} of the modules in the chain, and must be resolved, if at all, by the code between them (Fig.~\ref{fig:perception}a).

\finding{Semantic displacement.}
This is where a code-writing optimizer does something predictable and harmful.
It cannot change what a model understands, so it implements the missing semantics in the substrate it \emph{can} change: geometry.
The fridge region skill accumulated rules about recesses, occluding boundaries, adjacent thin regions and scene-scale edges, each with a tunable threshold, until it held 33 thresholds in about three thousand lines, composing an object localizer with 48 more.
We call this \emph{semantic displacement}: a semantic capability that the substrate lacks gets displaced into procedural code that approximates it on the states where the loop sees it fail.
Displacement has a recognizable signature, and every part of it appeared in our run:
\begin{itemize}
\item \emph{It fits states, not meaning.} The fridge region gap had two recorded states. Across 27 attempts, one seed was rescued every time and the other never. Rules that encode ``the top shelf'' for one viewpoint do not encode it for another.
\item \emph{It passes gates and then decays.} A top-shelf fix passed every gate in round~36 and deployed well, and the front reopened once programs and viewpoints drifted. Tests on the states where the rules were fit cannot detect this.
\item \emph{It grows monotonically.} Each new failure adds a rule; no failure removes one. Accepted changes concentrated in perception (Fig.~\ref{fig:campaign}c) partly because perception is the first link of every chain (\S\ref{sec:composition}), and partly because displaced semantics keep generating new failures to patch.
\end{itemize}
Displacement is not an agent error. Given the substrate, it is the best available move, and a stronger proposer would make it faster.

\finding{Seeing versus understanding.}
The problem is made harder because relational perception is entangled with viewpoint.
With the fridge door open and the robot beside it, the top shelf's own structure is invisible; only the items standing on it show, sometimes as a few hundred pixels and on one seed as almost nothing.
A system must therefore tell apart ``I cannot see the region'', which calls for moving, from ``I cannot tell which region'', which calls for better reasoning.
Displaced geometry cannot make this distinction: a rule tuned to answer ``top shelf'' will answer something whether or not the shelf is visible.
The evaluator had the same difficulty. For much of the run it could not tell these cases apart either, and early on its own language matching failed on ``refrigerator'' versus \code{fridge} and accepted a ``top shelf'' answer that landed on the lower shelf's front lip. Open-vocabulary semantics are hard for the judge as well as for the robot.

\finding{The escape is a change of model class, and it lay outside the loop.}
A 7B open-weights vision-language model (Qwen2.5-VL~\citep{bai2025qwen25vl}), queried directly on recorded head-camera frames, placed ``the top shelf'' and ``the shelf below'' correctly in all four recorded head views we tried, in about a second per query. Models of this class resolve relations over the whole scene rather than per mask~\citep{chen2024spatialvlm,yuan2024robopoint}.
The loop did not get there on its own. Its proposers kept asking for \emph{repairs} of the Florence-2 wrapper (``more region specificity'', ``more hypothesis diversity''), which is how an agent reasons when its task is to fix a failing component rather than to question the component's class. Many of those requests were silently lost (\S\ref{sec:learning}).
There is an irony here. The proposer is itself a large vision-language model, and it was shown camera keyframes of every failed top-shelf step (\S\ref{sec:system}). Models of its kind can tell which shelf is on top in such an image. What it could not do was give the robot that ability, because the only channel from the agent's understanding to the robot's behavior was code. The agent's own perception did not transfer to the robot's perception.
The vision-language class entered only after a human suggested it. Even then, the first agent-built grounders that passed research were registered under a family whose code never calls them, and so never ran on the robot.

\insight{1}{Open-instruction perception fails in a self-improving loop not because each module is weak but because the relation lives between modules, where only code can reach it. A code-writing agent then displaces the missing semantics into ever-growing geometry. The loop needs perception whose representation contains the relation (reason first, verify with geometry), and it needs model-class change to be an action the agent can take, not a repair it can request.}

\section{Lesson 2: skill chains lock learning onto the first step}
\label{sec:composition}

A household task is not a skill but a chain of them. The fridge task needs roughly fifteen declared steps: find the top shelf, clear it, then twice find a condiment, approach, grasp, carry, find the shelf again, place and release.
This section argues that chaining does more than multiply small unreliabilities. It also shapes \emph{what the loop can learn}, in a way that keeps it stuck at the front of the chain.

\finding{Double attenuation.}
Consider a change that raises the reliability of step $j$ by $\Delta p_j$ and leaves everything else fixed. Its effect on task success is
\begin{equation}
\Delta S \;=\; \underbrace{\pi_j}_{\text{reach}}\;\Delta p_j\;\underbrace{\textstyle\prod_{i>j}p_i}_{\text{continue}},
\label{eq:attenuation}
\end{equation}
so a gain is discounted both by the chance of reaching step $j$ and by the chance of finishing after it.
The same reach factor $\pi_j$ also discounts the \emph{evidence} about step $j$. Of $n$ rollouts or fresh test states, only about $n\pi_j$ exercise step $j$ at all. Downstream steps therefore produce few failures to learn from, and their gates see few states on which a candidate can differ from the incumbent.
A downstream repair is attenuated twice, once in its value and once in its measurability, by the same upstream unreliability.
We call the result \emph{frontier lock}: learning concentrates on the first failing link because only there is evidence plentiful and gates resolvable, and the rest of the chain stays under-trained no matter how many rounds run.

\begin{figure*}[t]
\centering\includegraphics[width=\textwidth]{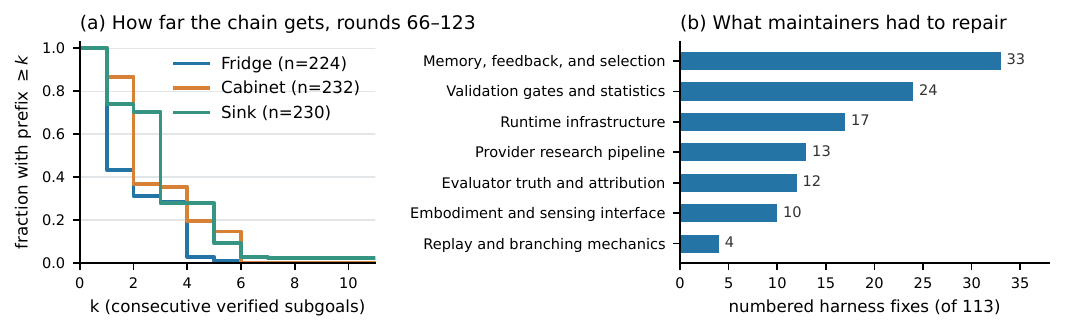}
\caption{\textbf{(a) Frontier lock.} Fraction of rollouts in the three-task phase whose first $k$ declared steps all passed. More than half of fridge rollouts fail the first step and almost none reach the fifth, so steps beyond the frontier generate little evidence and are rarely exercised by gates.
\textbf{(b) Where the human effort went.} The 113 harness repairs made by humans, grouped by what they repaired (the round each one was found in is listed in Table~\ref{tab:rounds}). Most concern what the agents were told, remembered and rewarded for, not the simulator or the replay machinery (\S\ref{sec:learning}).}
\label{fig:prefix}
\end{figure*}

Figure~\ref{fig:prefix}a shows the lock directly. More than half of the fridge rollouts failed the first step, and almost none got past the fourth. The perception front absorbed nearly all accepted changes (Fig.~\ref{fig:campaign}c) not because the agent preferred perception but because that is where the loop's evidence was.
The grasp shows the other side. When a grasp candidate was tested on fresh states, most of those states failed at localization or reach in both arms, so the grasp never ran. Several acceptance rules (an absolute success floor, a multi-task span requirement, a holdout requirement) were at some point impossible for \emph{any} grasp candidate to satisfy for this reason alone. A grasp change that did win reproduced paired comparisons on unseen states, including one on the held-out task, was followed by five rounds without a single task success. That is Eq.~\ref{eq:attenuation} observed: a real local gain multiplied by near-zero reach and continuation factors.

\finding{Failures surface downstream of their causes.}
In a chain, the step that fails is often not the step that is wrong. A localizer that returns a confident wrong point causes the reach, grasp and place that follow to fail, each for its own reason. A pre-grasp motion that stops short produces an evaluator verdict of ``no grasp contact'', which is literally true. A motion that cannot reach its target often means the base was parked in the wrong place two steps earlier.
Because the loop attributes each failure to the step where it surfaced, it sent proposers to repair the wrong family, sometimes for many rounds. This is the representability condition failing \emph{because of} composition, not because of any single evaluator.
The corrective is to judge each step by what it could have done from where it started. For example, the evaluator can run its own reach trial before judging a motion, and reassign the failure to placement when even that trial cannot get there.

\finding{Composition inside skills, and the chain nobody learns.}
Skills also call skills: region finding calls the object localizer, and the grasp calls motion and approach. A change to one can silently break another that composes it, and until late in the run the replay machinery could not branch at a nested call at all, so gaps rooted inside a composed skill were untestable.
Above the skills, the chain itself was not a learning target. The orchestrator wrote each program once and was told to route every failure into the skill library, not around it. It never learned to insert a missing step, such as raising the torso and looking again before grounding a shelf. Every chain-level problem therefore reappeared as some skill's gap, where it could not be fixed.

\finding{Task success is not program success.}
The easiest task, placing an item in the sink, is the only one that ever succeeded, and most of its successes were accidents. The program's own place steps had failed, the gripper opened elsewhere, and the object happened to fall into the sink. The same program alternated between success and failure across rounds because an upstream estimate took one of a few values depending on contact physics.
Native task success and correct execution of the intended chain are different measurements. A loop that counts the first as progress will reinforce luck.

\insight{2}{Chaining makes self-improvement path-dependent. Upstream unreliability discounts both the value and the measurability of every downstream repair, so the loop locks onto the frontier and never gathers evidence for the rest of the chain. Robot self-improvement must learn at the level of the chain: skills with explicit preconditions and recovery, attribution by what a step could have done from where it started, an orchestrator that can edit the chain itself, and test states drawn to exercise each step rather than only the first.}

\section{Lesson 3: what the agent learns is decided by the harness}
\label{sec:learning}

In a self-improvement loop the proposer supplies variation; everything else (the evaluators that say what went wrong, the gates that accept or reject, the memory that carries experience forward) supplies selection. In software the selection side is a test suite. In robotics it is a set of fallible models of physics and meaning, written by people and never complete.
This section argues that the selection side, not the proposer, determined what our loop learned. We identify four mechanisms by which it failed. Each one turns a capable optimizer into an efficient producer of the wrong thing.

\finding{Evaluator capture.}
A strong optimizer finds the evaluator's errors before it finds the task.
For much of the run, ``reachable'' was judged by a sphere around the arm base. The agent's approach skill learned to park the base exactly on the sphere's surface: hundreds of recorded successes lie within a millimeter of the advertised radius, and the grasps that followed could not reach. Meanwhile, the one physical mechanism that did make the top shelf reachable, a torso lift, was missing from the action description. Proposers reported dozens of times that they lacked ``the actuator layout''. The base controller's advertised speed was off by more than an order of magnitude and hid a dead zone in which every short move stalled.
None of this is reward hacking in the adversarial sense~\citep{amodei2016concrete}. The agent did exactly what it was told, and it was told wrong things.
The deeper point is that in a robot loop the evaluator and the embodiment description are \emph{models of the world}, not specifications. Their error is a second optimization target that the loop exploits by default, and the more capable the proposer, the faster it does so.

\finding{Unresolvable selection.}
A gate is a statistical test on a handful of physical rollouts, and ours failed in every direction a test can fail.
It was \emph{too coarse}: exact-state replay demanded a 60\% rescue rate on gaps that usually had one or two recorded states, so a design that rescued one of two was rejected by arithmetic, and the one gate that measures generalization on fresh states was unreachable for partial designs.
It was \emph{too permissive}: when fresh baselines were already saturated, the required improvement was zero, and about two dozen changes were accepted on replay of the very states the proposer had seen, which is precisely what an over-fitted patch passes.
It was \emph{noisier than the effects}: when disagreeing fresh-state pairs were re-run, nearly half did not reproduce. Contact physics, upstream estimates and provider behavior vary between runs even from a shared seed.
And it was \emph{blind} beyond the frontier (\S\ref{sec:composition}).
Every one of these was fixable, and was fixed. The general lesson is that resolvability must be \emph{designed}: thresholds that respect the sample size, reproduction of every difference, states drawn to exercise the changed skill, and a lockbox the loop has never selected against~\citep{dwork2015generalization,agarwal2021deep}.

\finding{The evidence fixed point.}
The simulator is deterministic, collection reused four seeds per task, and the orchestrator's memory froze its best programs. Together these made the loop's input constant. For more than a dozen consecutive rounds the fridge and cabinet rollouts were identical, the miner re-filed the same failures, and proposers answered gaps they had seen over twenty times, reading hundreds of thousands of tokens of their own history per attempt.
A loop whose observations do not change cannot learn anything new from them, however good its proposer. Repetition makes a useful regression sentinel; discovery needs a separate, deliberate source of novelty.

\finding{Memory corruption.}
Memory is what makes the loop recursive, and a corrupted memory is worse than none.
The proposer could ask for research on a missing capability. For much of the run, requests written to one of the two permitted locations were treated as library code: gated, found unable to move the robot, and recorded as \emph{failed attempts}. On the fridge-region front the same request, for a more specific region model, was filed round after round and never researched.
The proposer, reading its history, concluded the request was on file and declined the gap, writing ``I do not duplicate that request''.
An untested hypothesis had been recorded as a refuted one, and the record then suppressed the very exploration that could have escaped semantic displacement (\S\ref{sec:perception}).
Beliefs went stale in the same way: a proposer argued from its history that a repaired model was not being served, contradicting every trace of the current round.
The fix was to give memory a type system: every hypothesis is recorded as tested and rejected, untestable, failed on infrastructure, pending, or never run, and only the first counts against it.

\finding{Capability is not use.}
The research pipeline eventually turned requests into registered models at a steady rate, yet none of the eleven models first registered in the final rounds was ever called by the deployed robot. Registration showed that a model answers a synthetic probe and does not break what was tested. It did not show that any skill routes to it, and no part of the loop was responsible for integration.

\finding{Framing outweighs effort.}
Proposer behavior was far more sensitive to how the harness described its own gates than to how hard the proposer thought. A single paragraph warning that state-specific repairs are rejected coincided with proposers declining every gap for two rounds. A paragraph explaining that rejected attempts seed the next one coincided with nearly every attempt becoming a real edit, one of which was promoted. Raising the proposer's reasoning effort changed nothing visible.
The harness's text is part of the learning algorithm.

\finding{Who actually learned.}
Over the run, humans made 113 numbered repairs to the harness (Fig.~\ref{fig:prefix}b), about twice the number of library changes the agent had accepted. Most corrected what the agents were told, remembered or rewarded for. Several were first diagnosed by the agent itself: a proposer that declined a gap and explained why was often pointing at a harness bug, such as nested calls renumbered on replay or a program asking the object finder for ``cabinet interior''.
In this sense the loop was recursive in the wrong place. The agent improved the robot within a harness that only humans improved, while the agent's most accurate diagnoses concerned the harness.

\insight{3}{In robotics the selection side of self-improvement (evaluators, embodiment descriptions, gates, memory and their wording) is a model of the world, not a specification. A capable proposer exploits its errors, cannot learn past its resolution, stalls when its inputs stop changing, and is misled by its own history. Recursive self-improvement must therefore extend to the selection side, with its own tests. The agent's reasoned refusals are a ready source of harness bug reports.}

\section{What we would do differently}
\label{sec:design}
If we started again, we would build the loop differently. Below are our recommendations, grouped by which of the three conditions each one restores. They come from this experiment but were not themselves tested in it, so Table~\ref{tab:design} pairs each with an experiment that could prove it wrong.

\finding{Restoring representability: say why, where it started.}
\emph{(D1) Attribute failures by capability, not by location.} An evaluator should ask whether a step could have succeeded from the state in which it began, and assign the failure to the step that made it impossible. This counters the downstream surfacing of failures in chains.
\emph{(D2) Measure the embodiment instead of declaring it.} Actuator response, dead zones, reachable workspace and self-occlusion should be probed and published as data, and re-probed when they change. An agent can only exploit capabilities it is told about, and it will exploit errors it is told.
\emph{(D3) Feed back mechanism, not verdicts.} Rejected attempts should return execution traces and structured diagnostics. In our experiment this was the difference between designs that were re-derived from scratch and designs that converged.

\finding{Restoring expressibility: widen what the agent can change.}
\emph{(D4) Reason first, verify with geometry.} For open-vocabulary instructions, perception should include a model whose representation contains relations, such as a vision-language model producing points or boxes~\citep{bai2025qwen25vl,yuan2024robopoint,chen2024spatialvlm}. Segmentation~\citep{kirillov2023segment,carion2025sam3} and depth should convert its answer into geometry, and geometric checks should \emph{veto} rather than \emph{decide}. Perception must be able to abstain, and abstention should trigger looking.
\emph{(D5) Make model-class change an action.} The loop should detect semantic displacement (growing rule counts, rescues that hold on one recorded state but not its sibling, repeated repairs of one residual) and respond by researching a different class of model rather than repairing the current one.
\emph{(D6) Put the chain in the substrate.} The orchestrator should be allowed to edit the chain (insert a look-again step, reorder, add recovery), and such edits should be gated like skill changes. Otherwise chain-level problems can only be misfiled as skill gaps.

\finding{Restoring resolvability: test what the change could affect.}
\emph{(D7) Draw test states to exercise the changed step.} Counter frontier lock by sampling or constructing states that reach the changed step (for example by resetting the simulator past the frontier) instead of only running the task from the start.
\emph{(D8) Make gates statistical.} Apply rate thresholds only where the sample can resolve them, require every paired difference to reproduce, exclude cases the change could not influence, and keep a lockbox the loop never sees.
\emph{(D9) Separate sentinels from discovery.} Keep a fixed regression panel, but spend a declared share of each round on new states, layouts or viewpoints, so the loop's input keeps changing.

\finding{Closing the recursion.}
\emph{(D10) Type the memory.} Record every hypothesis with its test status, carry measured partial successes forward with their traces, and expire conclusions whose evaluator has changed.
\emph{(D11) Require use.} Admitting a model should create an integration obligation: a deployed skill must call it on relevant states before the admission counts as capability.
\emph{(D12) Improve the harness in the loop.} Treat evaluators, gates and harness text as learnable, with their own gate: replay the recorded archive under a proposed harness change and check which past decisions flip and why. Route the agent's reasoned refusals to this channel, since they were the most reliable harness bug reports we had.

\begin{table*}[t]
\centering\small
\caption{\textbf{Design implications and how to falsify them.} Every experiment should use frozen library snapshots, matched agent and simulator budgets, repeated seeds, and an evaluation set the loop has never selected against.}
\label{tab:design}
\renewcommand{\arraystretch}{1.12}
\begin{tabularx}{\textwidth}{@{}p{.24\textwidth}p{.12\textwidth}X@{}}
\toprule
Implication & Condition & Falsifying experiment\\
\midrule
D1 Attribute by capability & Repr. & Re-label the recorded archive with capability-based attribution; count gaps that move to a different family, and compare proposal success on relabelled vs.\ original gaps.\\
D2 Measured embodiment & Repr. & Run the loop with declared vs.\ probed actuator descriptions; compare how often candidates hit evaluator boundaries (e.g.\ values piling at an advertised constant).\\
D4 Reason, then verify & Expr. & Replace only the region grounder in a frozen library; measure region success and chain depth on new seeds and new kitchen layouts.\\
D5 Model-class escalation & Expr. & Repair-only proposer vs.\ proposer with an escalation policy at equal budget; rounds to the first fresh-state win on a relational front.\\
D6 Chain in the substrate & Expr. & Fixed programs vs.\ gated chain edits; per-step reach and conditional reliability, and task success.\\
D7 States past the frontier & Resol. & Task-start test states vs.\ states reset to the changed step; fraction of evaluable pairs and detection rate of known-good changes.\\
D8 Statistical gates & Resol. & Current vs.\ resolution-aware gates on the same candidate stream; rate of accepted changes that improve the lockbox.\\
D9 Novelty budget & Resol. & Fixed seeds vs.\ fixed panel plus new states; distinct failure mechanisms observed per round.\\
D10 Typed memory & Recursion & Replay proposers on the archive with true vs.\ mislabelled request status; re-proposal rate of useful requests.\\
D11 Integration obligation & Recursion & Registration alone vs.\ registration plus obligation; call rate on eligible states and paired task change.\\
D12 Harness in the loop & Recursion & Gate proposed harness changes by archive replay; fraction of later human fixes the loop would have found.\\
\bottomrule
\end{tabularx}
\end{table*}

\section{Related work}
\finding{Self-improving agents.}
Voyager grows an executable skill library with an automatic curriculum in Minecraft~\citep{wang2023voyager}. Eureka evolves reward code for robot learning~\citep{ma2023eureka}. Reflexion turns verbal feedback into better attempts~\citep{shinn2023reflexion}. The Darwin G\"odel Machine keeps an archive of self-modified coding agents under empirical selection~\citep{zhang2025darwin}.
In the terms of \S\ref{sec:framework}, these settings largely satisfy representability, expressibility and resolvability by construction: game state, rewards, unit tests or benchmark scores are exact, cheap, and aligned with the objective.

\finding{Agentic self-improvement for robots.}
Three recent systems are closest to ours.
RHO optimizes multi-file robot-policy repositories with tool-using coding agents and execution feedback~\citep{elmaaroufi2026rho}.
ASPIRE lets coding agents write and refine robot control programs in an open-ended loop~\citep{lu2026aspire}. A closed-loop execution engine exposes fine-grained multimodal traces for diagnosis and repair, validated fixes are distilled into a growing skill library, and evolutionary search explores task sequences and programs. It reports large gains on LIBERO-Pro, Robosuite and BEHAVIOR-1K.
ENPIRE builds a harness in which coding agents improve real-world robot policies~\citep{xiao2026enpire}. Its modules handle automatic environment reset and verification, policy training, parallel physical rollouts, and an evolution step in which agents read logs and literature and revise training code. With it, agents reach near-perfect success on dexterous real-world tasks.
Both results support our view that the agent is capable and that the harness decides what it can achieve. ENPIRE treats automatic reset and verification as a module of its own, and ASPIRE's engine is built to make failures diagnosable; these are the representability and resolvability conditions in our terms.
Our study complements them with a long, fully recorded run that did \emph{not} reach its goal: a fifteen-step household task with open-vocabulary relational instructions, where perception semantics, skill chains and a human-maintained harness became the limiting factors. We report why, not a benchmark gain.

\finding{Language-conditioned robot programs and grounding.}
Code as Policies~\citep{liang2023code}, SayCan~\citep{ahn2022saycan} and VoxPoser~\citep{huang2023voxposer} compose language models with perception and control, typically with hand-built perception.
Open-vocabulary segmentation~\citep{kirillov2023segment,carion2025sam3,xiao2024florence} and spatially grounded vision-language models~\citep{chen2024spatialvlm,yuan2024robopoint,bai2025qwen25vl} provide the building blocks.
Semantic displacement (\S\ref{sec:perception}) describes what happens when a code-writing improver has the former but not the latter: it rebuilds relational reasoning from geometry and does not converge.

\finding{Evaluation under selection.}
Reward misspecification~\citep{amodei2016concrete}, few-run statistical fragility~\citep{agarwal2021deep} and adaptive reuse of holdouts~\citep{dwork2015generalization} are well studied in isolation. In a self-improvement loop they compound: each accepted change is selected by the evaluator it may be exploiting, and becomes part of the environment for the next round.

\section{Limitations}
\label{sec:limitations}
This is one long run on one simulator, one kitchen layout and three tasks, with no independent replicates, matched-budget baselines or ablations. Harness repairs, model versions and providers changed along the same timeline, so our analysis identifies mechanisms from records and dissections, not causal effect sizes. We cannot say whether a different agent, more compute, or a different library representation would have solved the task.
The vision-language check in \S\ref{sec:perception} covers a handful of recorded views and shows feasibility, not benchmark performance.
The framework of \S\ref{sec:framework} is deliberately simple. Its value is diagnostic: it tells which part of a loop to examine when accepted changes stop moving the task. It is not a predictive model.
The design implications in \S\ref{sec:design} are hypotheses, and Table~\ref{tab:design} states how each could be refuted.
The appendix gives a round-by-round record of the run (Table~\ref{tab:rounds}) and the measurements behind each statement in the main text (Table~\ref{tab:numbers}).

\section{Conclusion}
A coding agent given a robot, a simulator and a rule against human-written skills did discover real capabilities, and it was a sharper diagnostician of its own harness than we expected. It did not solve a household task, and the reasons were not a shortage of code or of reasoning.
Relational meaning lived between perception modules, where only brittle code could reach it. Chains locked learning onto their first link. And the selection side of the loop, its evaluators, gates and memory, was a fallible model of the world that the agent exploited, and that only humans could improve.
Self-improvement in software rests on representable failures, expressible fixes and resolvable tests. In robotics these must be built, and we argue that building them, not stronger proposers, is where progress will come from.
Until then, the number of accepted changes measures activity, not progress.

\FloatBarrier
\begingroup\small
\bibliographystyle{unsrtnat}
\bibliography{references}
\endgroup
\appendix
\onecolumn
\section{Round-by-round record}
\label{app:rounds}
Table~\ref{tab:rounds} lists every round: how the tasks went, which problems the proposer worked on, what it proposed and how those proposals ended, which skill changes were accepted, which external models were registered, and which harness problems the round exposed. It shows the three phases of \S\ref{sec:results} in detail. Early rounds make and accept changes on the fridge task. The middle rounds turn up harness problems one after another. The late rounds keep working on the same four problems (find object, find region, grasp, reach) while registering new models, without changing the outcome.

\begingroup\scriptsize
\setlength{\tabcolsep}{2.6pt}
\renewcommand{\arraystretch}{1.08}
\begin{longtable}{@{}r >{\raggedright}p{.07\textwidth} >{\raggedright}p{.17\textwidth} >{\raggedright}p{.12\textwidth} >{\raggedright}p{.2\textwidth} >{\raggedright}p{.16\textwidth} >{\raggedright\arraybackslash}p{.19\textwidth}@{}}
\caption{\textbf{Round-by-round record.} \emph{Outcome}: task successes / completed rollouts for \fridge\ (F), \cabinet\ (C) and \sink\ (S). \emph{Problems worked on}: the gaps sent to the proposer, as step: diagnosis. \emph{Proposals}: attempts, of which research requests (req.), declines (no change) and rejections by test (replay = exact failure replay; fresh = paired fresh-scene and held-out test; transfer = sealed recorded failures; regr/inv = regression and invariance checks). \emph{Accepted}: skill changes that passed every test, with the problem they fixed. \emph{Models}: external models registered by the research agent, and research jobs that did not register. \emph{Harness problems found}: harness repairs made by humans after this round exposed the problem.}\label{tab:rounds}\\
\toprule
Rd & Outcome & Problems worked on & Proposals & Accepted skill changes & Models & Harness problems found\\
\midrule
\endfirsthead
\toprule
Rd & Outcome & Problems worked on & Proposals & Accepted skill changes & Models & Harness problems found\\
\midrule
\endhead
\bottomrule
\endfoot
1 & F\,0/4 & find obj: wrong object & 1 att.; rej.\ replay 1 & -- & -- & --\\
2 & F\,0/4 & find obj: no task object near point & 2 att.; rej.\ replay 2 & -- & -- & target\_\allowbreak{}not\_\allowbreak{}in\_\allowbreak{}view attributed on visibility\\
3 & F\,0/4 & find obj: no task object near point & 2 att.; rej.\ replay 2 & -- & -- & effect packs visible through worker wrappers\\
4 & F\,0/4 & find obj: target not in view & 2 att.; 1 req.; rej.\ replay 1 & -- & 1 not reg. & agent-extensible provider kinds\\
5 & F\,0/4 & find obj: target not in view & 2 att.; 1 req.; rej.\ replay 1 & -- & target\_\allowbreak{}free\_\allowbreak{}active\_\allowbreak{}view\_\allowbreak{}base\_\allowbreak{}placement & --\\
6 & F\,0/4 & find region: wrong fixture & 2 att.; rej.\ replay 1, fresh 1 & -- & -- & --\\
7 & F\,0/4 & find obj: target not in view; find region: wrong fixture & 4 att.; rej.\ replay 4 & -- & -- & full diffs kept in candidate memory\\
8 & F\,0/4 & find obj: target not in view; find region: wrong fixture & 4 att.; rej.\ replay 4 & -- & -- & counterfactual gate semantics documented\\
9 & F\,0/4 & find obj: target not in view; find region: wrong fixture & 4 att.; rej.\ replay 4 & -- & -- & trial execution traces fed back\\
10 & F\,0/4 & find obj: target not in view; find region: wrong fixture & 3 att.; rej.\ replay 2 & fixture\_\allowbreak{}region routing, fixture\_\allowbreak{}region.\allowbreak{}deployed\_\allowbreak{}localize (find region: wrong fixture) & -- & --\\
11 & F\,0/4 & find obj: target not in view; find region: wrong fixture & 3 att.; 1 no change; rej.\ replay 2 & -- & -- & --\\
12 & F\,0/4 & find obj: target not in view; find region: wrong fixture & 4 att.; 1 no change; rej.\ fresh 2, replay 1 & -- & -- & --\\
13 & F\,0/4 & find obj: target not in view; find region: fixture not matched & 6 att.; rej.\ replay 6 & -- & -- & elitism seeding\\
14 & F\,0/4 & find obj: target not in view; find region: fixture not matched & 6 att.; 1 req.; rej.\ replay 3, fresh 2 & -- & equivariant\_\allowbreak{}grid\_\allowbreak{}base\_\allowbreak{}placement & per-case fresh-state feedback\\
15 & F\,0/4 & find obj: target not in view; find region: fixture not matched & 5 att.; 1 req.; rej.\ fresh 2, replay 1 & localize.\allowbreak{}calibrated\_\allowbreak{}multiview\_\allowbreak{}rgb (find obj: target not in view) & 1 not reg. & --\\
16 & F\,0/4 & find obj: target not in view; find region: fixture not matched & 4 att.; 1 no change; rej.\ replay 3 & -- & -- & --\\
17 & F\,0/4 & find obj: target not in view; find region: fixture not matched; move: target not reached & 9 att.; 1 req.; rej.\ replay 4, error 3 & localize.\allowbreak{}calibrated\_\allowbreak{}multiview\_\allowbreak{}rgb (find obj: target not in view) & 1 not reg. & era-aware selective memory\\
18 & F\,0/4 & find obj: no task object near point; find region: fixture not matched; move: target not reached & 7 att.; 1 req.; 1 no change; rej.\ replay 5 & -- & target\_\allowbreak{}graspgen\_\allowbreak{}6dof & --\\
19 & F\,0/4 & find obj: target not in view; find region: fixture not matched; move: target not reached & 7 att.; 1 req.; rej.\ replay 5 & localize.\allowbreak{}calibrated\_\allowbreak{}multiview\_\allowbreak{}rgb (find obj: target not in view) & relational\_\allowbreak{}rgbd\_\allowbreak{}localizer & --\\
20 & F\,0/4 & find obj: target not found; find region: wrong fixture; move: target not reached & 5 att.; rej.\ replay 3 & fixture\_\allowbreak{}region.\allowbreak{}deployed\_\allowbreak{}localize (find region: wrong fixture); localize.\allowbreak{}sam3\_\allowbreak{}1\_\allowbreak{}da3 (find obj: target not found) & -- & --\\
21 & F\,0/4 & find obj: target not found; find region: fixture not matched; move: target not reached & 6 att.; 1 req.; 1 no change; rej.\ replay 4 & -- & mplib\_\allowbreak{}whole\_\allowbreak{}body\_\allowbreak{}goalset & --\\
22 & F\,0/4 & find obj: target not found; find region: fixture not matched; move: target not reached & 6 att.; 1 req.; 1 no change; rej.\ replay 4 & -- & relational\_\allowbreak{}rgbd\_\allowbreak{}localizer & --\\
23 & F\,0/4 & find obj: target not in view; find region: fixture not matched; move: target not reached & 7 att.; 1 req.; rej.\ replay 5, fresh 1 & -- & mplib\_\allowbreak{}confined\_\allowbreak{}motion & --\\
24 & F\,0/4 & find obj: program exception; find region: fixture not matched; move: target not reached & 9 att.; rej.\ replay 9 & -- & -- & fixture-region evaluator truthful (refrigerator/fridge, 3-D containment)\\
25 & F\,0/4 & -- & -- & -- & -- & --\\
26 & F\,0/4 & find obj: no task object near point; find region: wrong fixture; move: target not reached & 6 att.; 1 req.; rej.\ replay 3, fresh 1 & localize.\allowbreak{}calibrated\_\allowbreak{}multiview\_\allowbreak{}rgb (find obj: no task object near point) & equivariant\_\allowbreak{}grid\_\allowbreak{}base\_\allowbreak{}placement & --\\
27 & F\,0/4 & find obj: no task object near point; find region: wrong region; move: target not reached & 6 att.; 1 req.; 1 no change; rej.\ replay 4 & -- & equivariant\_\allowbreak{}grid\_\allowbreak{}base\_\allowbreak{}placement & evaluator change bounds prior-attempt era\\
28 & F\,0/4 & find obj: no task object near point; find region: wrong region; move: target not reached & 4 att.; 2 req.; 1 no change; rej.\ replay 1 & -- & 2 not reg. & materialization failures enter refinement; probe synthesizes proprio/control inputs\\
29 & F\,0/4 & find obj: no task object near point; find region: wrong region; move: target not reached & 7 att.; 1 req.; 1 no change; rej.\ replay 3, fresh 2 & -- & relational\_\allowbreak{}rgbd\_\allowbreak{}localizer & --\\
30 & F\,0/4 & find obj: no task object near point; find region: wrong region; move: target not reached & 4 att.; 1 req.; 1 no change; rej.\ replay 1 & localize routing, localize.\allowbreak{}calibrated\_\allowbreak{}multiview\_\allowbreak{}rgb (find obj: no task object near point) & relational\_\allowbreak{}rgbd\_\allowbreak{}localizer & provider repairs do not fork providers\\
31 & F\,0/4 & find obj: program exception; find region: wrong region; move: target not reached & 7 att.; 1 no change; rej.\ replay 6 & -- & -- & --\\
32 & F\,0/4 & find obj: program exception; find region: wrong region; move: target not reached & 7 att.; 1 no change; rej.\ replay 5, fresh 1 & -- & -- & --\\
33 & F\,0/4 & find obj: program exception; find region: wrong region; move: target not reached & 7 att.; 1 no change; rej.\ replay 6 & -- & -- & baseline replay control; unreplayable segments retire\\
34 & F\,0/4 & find obj: program exception; find region: wrong region; move: target not reached & 6 att.; 1 req.; 2 no change; rej.\ replay 3 & -- & relational\_\allowbreak{}rgbd\_\allowbreak{}localizer & --\\
35 & F\,0/4 & find obj: program exception; find region: wrong region; move: target not reached & 5 att.; 2 no change; rej.\ replay 3 & -- & -- & --\\
36 & F\,0/4 & find obj: program exception; find region: wrong region; move: target not reached & 3 att.; 2 no change & fixture\_\allowbreak{}region.\allowbreak{}deployed\_\allowbreak{}localize (find region: wrong region) & -- & --\\
37 & F\,0/4 & find obj: target not in view; find region: wrong region; move: target not reached & 7 att.; 1 no change; rej.\ replay 6 & -- & -- & controller-grounded reachable evaluator; real actuator surface (torso, layout, IK); upstream evaluator changes expire evidence\\
38 & F\,0/4 & find obj: target not in view; find region: wrong region; move: target not reached & 5 att.; 1 req.; 1 no change; rej.\ fresh 3 & -- & relational\_\allowbreak{}rgbd\_\allowbreak{}localizer & --\\
39 & F\,0/4 & find obj: target not in view; find region: wrong region & 9 att.; rej.\ replay 5, fresh 3 & localize.\allowbreak{}calibrated\_\allowbreak{}multiview\_\allowbreak{}rgb (find obj: target not in view) & -- & --\\
40 & F\,0/4 & find obj: target not in view; find region: wrong region & 3 att.; 1 no change & fixture\_\allowbreak{}region.\allowbreak{}deployed\_\allowbreak{}localize (find region: wrong region); localize.\allowbreak{}calibrated\_\allowbreak{}multiview\_\allowbreak{}rgb (find obj: target not in view) & -- & --\\
41 & F\,0/4 & find obj: target not in view; find obj: wrong object; find region: wrong region & 3 att.; 1 no change & localize.\allowbreak{}calibrated\_\allowbreak{}multiview\_\allowbreak{}rgb (find obj: target not in view); localize.\allowbreak{}calibrated\_\allowbreak{}multiview\_\allowbreak{}rgb (find obj: wrong object) & -- & --\\
42 & F\,0/4 & find obj: no task object near point; find obj: wrong object; find region: wrong region & 3 att.; 1 no change & localize.\allowbreak{}sam3\_\allowbreak{}1\_\allowbreak{}da3 (find obj: no task object near point); localize routing, localize.\allowbreak{}calibrated\_\allowbreak{}multiview\_\allowbreak{}rgb (find obj: wrong object) & -- & --\\
43 & F\,0/4 & find region: wrong fixture; find region: wrong region; reach: target out of workspace & 7 att.; 1 req.; 1 no change; rej.\ fresh 2, replay 2 & fixture\_\allowbreak{}region routing, fixture\_\allowbreak{}region.\allowbreak{}deployed\_\allowbreak{}localize (find region: wrong region) & mplib\_\allowbreak{}free\_\allowbreak{}space\_\allowbreak{}reachability & --\\
44 & F\,0/4 & find obj: no task object near point; find obj: target not in view; find region: wrong region & 5 att.; 1 no change; rej.\ replay 3 & localize routing, localize.\allowbreak{}calibrated\_\allowbreak{}multiview\_\allowbreak{}rgb (find obj: no task object near point) & -- & episode budget sized to task horizon; aborted subgoals keep evaluator residual\\
45 & F\,0/4 & find obj: no task object near point; find obj: target not in view; find region: wrong region & 4 att.; 2 no change; rej.\ replay 1 & localize.\allowbreak{}calibrated\_\allowbreak{}multiview\_\allowbreak{}rgb (find obj: no task object near point) & -- & --\\
46 & F\,0/4 & find region: wrong region; reach: target out of workspace & 5 att.; 1 req.; rej.\ replay 2 & fixture\_\allowbreak{}region.\allowbreak{}deployed\_\allowbreak{}localize (find region: wrong region); fixture\_\allowbreak{}region routing, fixture\_\allowbreak{}region.\allowbreak{}deployed\_\allowbreak{}localize (find region: wrong region) & mplib\_\allowbreak{}free\_\allowbreak{}space\_\allowbreak{}reachability & step cost is evidence\\
47 & F\,0/4 & find obj: no task object near point; find region: wrong region; reach: program exception & 5 att.; 1 req.; rej.\ error 1, replay 1 & fixture\_\allowbreak{}region.\allowbreak{}deployed\_\allowbreak{}localize (find region: wrong region); localize.\allowbreak{}calibrated\_\allowbreak{}multiview\_\allowbreak{}rgb (find obj: no task object near point) & mplib\_\allowbreak{}free\_\allowbreak{}space\_\allowbreak{}reachability & codex executable fallback\\
48 & F\,0/4 & find obj: no task object near point; find obj: target not in view; find region: fixture not found & 8 att.; rej.\ replay 7 & localize.\allowbreak{}sam3\_\allowbreak{}1\_\allowbreak{}da3 (find obj: no task object near point) & -- & program wall clock scales with step budget; unattributed task failure triggers revision\\
49 & F\,0/4 & find obj: target not in view; find region: fixture not found; find region: wrong region & 9 att.; rej.\ replay 9 & -- & -- & target geometry attached to not-in-view verdicts\\
50 & F\,0/4 & find obj: target not in view; find region: fixture not found; find region: wrong region & 7 att.; rej.\ error 4, replay 2 & localize routing, localize.\allowbreak{}calibrated\_\allowbreak{}multiview\_\allowbreak{}rgb (find obj: target not in view) & -- & --\\
51 & F\,0/4 & find obj: target not in view; find obj: wrong object & 9 att.; rej.\ error 9 & -- & -- & --\\
52 & F\,0/4 & find obj: target not in view; reach: target out of workspace & 6 att.; 1 req.; rej.\ replay 4 & localize.\allowbreak{}calibrated\_\allowbreak{}multiview\_\allowbreak{}rgb (find obj: target not in view) & mplib\_\allowbreak{}free\_\allowbreak{}space\_\allowbreak{}reachability & --\\
53 & F\,0/4 & find obj: no task object near point; find obj: target not found; find region: wrong region & 5 att.; rej.\ replay 3; infra/unrepl.\ 1 & fixture\_\allowbreak{}region.\allowbreak{}deployed\_\allowbreak{}localize (find region: wrong region) & -- & mining bounded to recent library generations\\
54 & F\,0/4 & find region: fixture not found; find region: wrong region; reach: target out of workspace & 4 att.; 1 req.; 1 no change; rej.\ replay 1; infra/unrepl.\ 1 & -- & mplib\_\allowbreak{}free\_\allowbreak{}space\_\allowbreak{}reachability & raising calls are branch targets\\
55 & F\,0/4 & find obj: target not in view; find obj: wrong object & 5 att.; rej.\ replay 2 & localize.\allowbreak{}calibrated\_\allowbreak{}multiview\_\allowbreak{}rgb (find obj: target not in view); localize.\allowbreak{}calibrated\_\allowbreak{}multiview\_\allowbreak{}rgb (find obj: wrong object); localize.\allowbreak{}calibrated\_\allowbreak{}multiview\_\allowbreak{}rgb (find obj: wrong object) & -- & structured diagnostics reach proposer\\
56 & F\,0/5 & find obj: no task object near point; find obj: target not in view; reach: target out of workspace & 6 att.; 1 req.; rej.\ replay 4 & localize.\allowbreak{}calibrated\_\allowbreak{}multiview\_\allowbreak{}rgb (find obj: no task object near point) & mplib\_\allowbreak{}free\_\allowbreak{}space\_\allowbreak{}reachability & base servo dead zone; truthful speed model\\
57 & F\,0/4 & find obj: target not in view; find region: no fixture near point; reach: target out of workspace & 7 att.; rej.\ replay 6 & fixture\_\allowbreak{}region.\allowbreak{}deployed\_\allowbreak{}localize (find region: no fixture near point) & -- & torso servo; truthful lift model\\
58 & F\,0/4 & find obj: target not found; find region: fixture not found; reach: target out of workspace & 7 att.; rej.\ replay 6 & localize routing, localize.\allowbreak{}calibrated\_\allowbreak{}multiview\_\allowbreak{}rgb (find obj: target not found) & -- & base stall watchdog; contact kind reported\\
59 & F\,0/4 & find obj: wrong object; grasp: no grasp contact; reach: target out of workspace & 9 att.; rej.\ replay 7, fresh 2 & -- & -- & native action layout advertises measured response; concurrent gap pipelines; effect gate cannot fail an unexercised family\\
60 & F\,0/4 C\,0/4 & find obj: target not in view; find region: fixture not found; grasp: no grasp contact; reach: target out of workspace & 10 att.; rej.\ replay 5, fresh 4 & fixture\_\allowbreak{}region.\allowbreak{}deployed\_\allowbreak{}localize (find region: fixture not found) & -- & --\\
61 & F\,0/4 C\,0/4 & find obj: wrong object; find region: fixture not found; grasp: closed above object; reach: target out of workspace & 12 att.; rej.\ replay 9, fresh 3 & -- & -- & elitism seeds from best partial rescue; strict paired improvement waives absolute floor\\
62 & F\,0/4 C\,0/4 & find obj: wrong object; find region: fixture not found; grasp: closed above object; reach: target out of workspace & 12 att.; rej.\ fresh 7, replay 4 & approach.\allowbreak{}metric\_\allowbreak{}reach\_\allowbreak{}plan (reach: target out of workspace) & -- & --\\
63 & F\,0/5 C\,0/4 & find obj: wrong object; find region: fixture not found; grasp: closed above object & 10 att.; rej.\ replay 6, fresh 2, regr 1 & grasp routing, grasp.\allowbreak{}analytic\_\allowbreak{}top (grasp: closed above object) & -- & orchestrator program memory\\
64 & F\,0/4 C\,0/4 & find obj: no task object near point; find region: no fixture near point; reach: target out of workspace & 10 att.; rej.\ replay 5, fresh 1, stale 1, regr 1 & fixture\_\allowbreak{}region.\allowbreak{}deployed\_\allowbreak{}localize, fixture\_\allowbreak{}region.\allowbreak{}metric\_\allowbreak{}masks (find region: no fixture near point); approach.\allowbreak{}metric\_\allowbreak{}reach\_\allowbreak{}plan (reach: target out of workspace) & -- & --\\
65 & F\,0/4 C\,0/4 & find obj: no task object near point; find region: no fixture near point; find region: wrong region; reach: target out of workspace & 10 att.; 1 req.; rej.\ replay 3, stale 1, error 1, fresh 1 & fixture\_\allowbreak{}region.\allowbreak{}deployed\_\allowbreak{}localize (find region: no fixture near point); fixture\_\allowbreak{}region.\allowbreak{}deployed\_\allowbreak{}localize (find region: wrong region); localize.\allowbreak{}calibrated\_\allowbreak{}multiview\_\allowbreak{}rgb (find obj: no task object near point) & mplib\_\allowbreak{}free\_\allowbreak{}space\_\allowbreak{}reachability & provider process recovery; unmeasured pairs excluded; fallback labels; gate baselines follow live library; three support tasks; sealed holdout task; program memory keyed by archived source; honest routing-fit partitions; reach-trial budget\\
66 & F\,0/1 C\,0/3 S\,0/2 & find obj: wrong object; find region: no fixture near point; grasp: no grasp contact; reach: target out of workspace & 9 att.; rej.\ replay 3, fresh 2; infra/unrepl.\ 3 & fixture\_\allowbreak{}region routing, fixture\_\allowbreak{}region.\allowbreak{}deployed\_\allowbreak{}localize (find region: no fixture near point) & mplib\_\allowbreak{}free\_\allowbreak{}space\_\allowbreak{}reachability & provider health shown to proposer; harness-filed provider repairs\\
67 & F\,0/4 C\,0/4 S\,0/4 & find obj: target not in view; find obj: wrong object; grasp: no grasp contact; reach: target out of workspace & 8 att.; 1 req.; rej.\ replay 4, fresh 1 & localize.\allowbreak{}calibrated\_\allowbreak{}multiview\_\allowbreak{}rgb (find obj: target not in view); approach.\allowbreak{}hold (reach: target out of workspace); routing refit & relational\_\allowbreak{}rgbd\_\allowbreak{}localizer & --\\
68 & F\,0/4 C\,0/4 S\,0/4 & find obj: target not in view; find region: wrong region; grasp: closed beside object; reach: target out of workspace & 11 att.; 1 req.; rej.\ replay 8, fresh 2 & routing refit & target\_\allowbreak{}graspgen\_\allowbreak{}6dof & numpy scalar hashing; fake simulator exposes torso\\
69 & F\,0/4 C\,0/4 S\,0/4 & find obj: target not in view; find region: wrong region; grasp: closed beside object; reach: target out of workspace & 12 att.; rej.\ replay 8, fresh 4 & routing refit & -- & generalization gap needs enough holdout cases\\
70 & F\,0/4 C\,0/4 S\,0/4 & find obj: target not in view; find region: fixture not found; grasp: closed above object; reach: target out of workspace & 12 att.; 1 req.; rej.\ replay 9, fresh 2 & routing refit & mplib\_\allowbreak{}confined\_\allowbreak{}motion & --\\
71 & F\,0/4 C\,0/4 S\,0/4 & find obj: target not in view; find region: fixture not found; grasp: closed beside object; reach: target out of workspace & 9 att.; rej.\ replay 6, fresh 1 & fixture\_\allowbreak{}region.\allowbreak{}deployed\_\allowbreak{}localize (find region: fixture not found); localize.\allowbreak{}calibrated\_\allowbreak{}multiview\_\allowbreak{}rgb (find obj: target not in view); routing refit & -- & --\\
72 & F\,0/4 C\,0/4 S\,0/4 & find obj: target not in view; find region: no fixture near point; grasp: closed above object; reach: target out of workspace & 12 att.; rej.\ replay 10, fresh 1 & fixture\_\allowbreak{}region.\allowbreak{}deployed\_\allowbreak{}localize (find region: no fixture near point); routing refit & -- & --\\
73 & F\,0/4 C\,0/4 S\,0/4 & find obj: no task object near point; find region: no fixture near point; grasp: closed above object; reach: target out of workspace & 9 att.; 1 req.; rej.\ replay 5 & fixture\_\allowbreak{}region.\allowbreak{}deployed\_\allowbreak{}localize (find region: no fixture near point); localize.\allowbreak{}calibrated\_\allowbreak{}multiview\_\allowbreak{}rgb (find obj: no task object near point); approach.\allowbreak{}hold, approach.\allowbreak{}metric\_\allowbreak{}reach\_\allowbreak{}plan (grasp: closed above object); routing refit & mplib\_\allowbreak{}free\_\allowbreak{}space\_\allowbreak{}reachability & view precondition judged at block start; recorded rescues outrank precondition ownership; reach failures split by mechanism; empty provider answers count as starvation; agent container isolation; transfer gate and unseen-win rule; sensor.\allowbreak{}depth not privileged\\
74 & F\,0/4 C\,0/4 S\,0/4 & find obj: wrong object; find region: wrong fixture; move: target not reached; reach: tool path blocked & 10 att.; 1 req.; rej.\ replay 5, error 1, transfer 1, fresh 1; infra/unrepl.\ 1 & routing refit & mplib\_\allowbreak{}whole\_\allowbreak{}body\_\allowbreak{}goalset, relational\_\allowbreak{}rgbd\_\allowbreak{}localizer; 1 not reg. & --\\
75 & F\,0/4 C\,0/4 S\,0/4 & find obj: target not in view; find region: wrong fixture; move: target not reached; reach: tool path blocked & 11 att.; 1 req.; rej.\ replay 5, fresh 3, transfer 1 & fixture\_\allowbreak{}region.\allowbreak{}deployed\_\allowbreak{}localize (find region: wrong fixture); routing refit & equivariant\_\allowbreak{}grid\_\allowbreak{}base\_\allowbreak{}placement, relational\_\allowbreak{}rgbd\_\allowbreak{}localizer; 1 not reg. & generalization gap compares like residuals; second fresh draw; side-effects gate; repairs replayed on recorded inputs; self-filtered depth\\
76 & F\,0/4 C\,0/4 S\,0/4 & find obj: wrong object; find region: wrong region; move: target not reached; reach: program exception & 7 att.; 2 no change; rej.\ replay 4 & fixture\_\allowbreak{}region.\allowbreak{}deployed\_\allowbreak{}localize (find region: wrong region); routing refit & mplib\_\allowbreak{}free\_\allowbreak{}space\_\allowbreak{}reachability, mplib\_\allowbreak{}whole\_\allowbreak{}body\_\allowbreak{}goalset & program faults not mining evidence; provider time excluded from program wall budget\\
77 & F\,0/4 C\,0/4 S\,0/4 & find obj: wrong object; find region: wrong region; move: target not reached; reach: tool path blocked & 7 att.; 3 no change; rej.\ replay 4 & routing refit & loftr\_\allowbreak{}multiview\_\allowbreak{}localization, mplib\_\allowbreak{}whole\_\allowbreak{}body\_\allowbreak{}goalset & tool-mount probe retried; motion failures split precondition vs motion; masked implementation names requirement; replay gate sample size\\
78 & F\,0/4 C\,0/4 S\,2/4 & find obj: target not in view; find region: wrong region; move: target not reached; reach: tool path blocked & 8 att.; 2 no change; rej.\ replay 6 & routing refit & mplib\_\allowbreak{}whole\_\allowbreak{}body\_\allowbreak{}goalset; 1 not reg. & --\\
79 & F\,0/4 C\,0/4 S\,2/4 & find obj: wrong object; find region: wrong region; move: target not reached; reach: tool path blocked & 8 att.; 2 no change; rej.\ replay 6 & routing refit & 1 not reg. & rotate away from declined kinds; routing credibility scales with gap size\\
80 & F\,0/4 C\,0/4 S\,1/4 & find obj: wrong object; find region: wrong region; grasp: localization offset; move: tool path blocked & 8 att.; 2 no change; rej.\ replay 6 & routing refit & 1 not reg. & --\\
81 & F\,0/4 C\,0/4 S\,1/4 & find obj: target not in view; find region: wrong region; grasp: closed beside object; reach: tool path blocked & 8 att.; 2 no change; rej.\ replay 4, regr 1, fresh 1 & routing refit & loftr\_\allowbreak{}multiview\_\allowbreak{}localization; 1 not reg. & keep successful programs\\
82 & F\,0/4 C\,0/4 S\,2/4 & find obj: target not in view; find region: no fixture near point; find region: wrong region; move: tool path blocked & 10 att.; 1 no change; rej.\ replay 9 & routing refit & 2 not reg. & --\\
83 & F\,0/4 C\,0/4 S\,1/4 & find obj: wrong object; find region: wrong region; move: target not reached; reach: tool path blocked & 8 att.; 2 no change; rej.\ replay 5, fresh 1 & routing refit & 1 not reg. & --\\
84 & F\,0/4 C\,0/4 S\,0/4 & find obj: target not in view; find region: no fixture near point; find region: wrong region; grasp: localization offset & 10 att.; 1 no change; rej.\ replay 9 & routing refit & loftr\_\allowbreak{}multiview\_\allowbreak{}localization & routing changes effect-gated; repair cooldown\\
85 & F\,0/4 C\,0/4 S\,0/4 & find obj: target not in view; find region: wrong region; move: target not reached; reach: tool path blocked & 8 att.; 2 no change; rej.\ replay 6 & -- & mplib\_\allowbreak{}free\_\allowbreak{}space\_\allowbreak{}reachability & --\\
86 & F\,0/4 C\,0/4 S\,0/4 & find obj: target not in view; find region: no fixture near point; find region: wrong region; move: tool path blocked & 10 att.; 1 no change; rej.\ replay 9 & routing refit & -- & stalled kinds rotate out\\
87 & F\,0/4 C\,0/4 S\,0/4 & find obj: target not in view; find region: wrong fixture; grasp: closed beside object; reach: tool path blocked & 8 att.; 2 no change; rej.\ replay 6 & routing refit & loftr\_\allowbreak{}multiview\_\allowbreak{}localization & --\\
88 & F\,0/4 C\,0/4 S\,0/4 & find obj: target not in view; find region: no fixture near point; find region: wrong region; move: tool path blocked & 10 att.; 1 no change; rej.\ replay 8, fresh 1 & routing refit & 1 not reg. & replay early stop; rejected repairs cool down\\
89 & F\,0/4 C\,0/4 S\,0/4 & find obj: wrong object; find region: wrong fixture; find region: wrong region; reach: tool path blocked & 8 att.; 1 no change; rej.\ replay 6 & localize routing, localize.\allowbreak{}calibrated\_\allowbreak{}multiview\_\allowbreak{}rgb (find obj: wrong object); routing refit & mplib\_\allowbreak{}whole\_\allowbreak{}body\_\allowbreak{}goalset; 1 not reg. & provider registration effect-gated\\
90 & F\,0/4 C\,0/4 S\,0/4 & find obj: target not in view; find region: wrong region; grasp: no grasp contact; move: target not reached & 7 att.; 3 no change; rej.\ replay 4 & -- & loftr\_\allowbreak{}multiview\_\allowbreak{}localization & composing families effect-gated\\
91 & F\,0/4 C\,0/4 S\,0/4 & find obj: wrong object; find region: wrong fixture; grasp: localization offset; reach: tool path blocked & 6 att.; 3 no change; rej.\ replay 3 & routing refit & -- & --\\
92 & F\,0/4 C\,0/4 S\,0/4 & find obj: target not in view; find region: wrong fixture; find region: wrong region; grasp: closed above object & 8 att.; 2 no change; rej.\ replay 6 & routing refit & -- & region visibility judged at block start\\
93 & F\,0/4 C\,0/4 S\,1/4 & find obj: wrong object; grasp: localization offset; reach: tool path blocked & 4 att.; 4 no change & routing refit & 1 not reg. & --\\
94 & F\,0/4 C\,0/4 S\,0/4 & find obj: target not in view; find region: wrong region; grasp: no grasp contact; move: target not reached & 5 att.; 4 no change; rej.\ replay 1 & routing refit & -- & proposer prompted for evidence-backed attempts\\
95 & F\,0/4 C\,0/4 S\,0/4 & find region: no fixture near point; find region: wrong fixture; grasp: localization offset; reach: tool path blocked & 8 att.; 4 no change; rej.\ replay 4 & routing refit & -- & --\\
96 & F\,0/4 C\,0/4 S\,0/4 & find obj: target not in view; find obj: wrong object; find region: wrong region; grasp: no grasp contact & 12 att.; 1 no change; rej.\ replay 9, inv 1 & localize.\allowbreak{}calibrated\_\allowbreak{}multiview\_\allowbreak{}rgb (find obj: wrong object); routing refit & 2 not reg. & --\\
97 & F\,0/4 C\,0/4 S\,0/4 & find obj: target not in view; find region: wrong fixture; grasp: no grasp contact; reach: tool path blocked & 10 att.; 2 no change; rej.\ replay 6, regr 1, fresh 1 & routing refit & 1 not reg. & --\\
98 & F\,0/4 C\,0/4 S\,1/4 & find obj: target not in view; find region: wrong region; grasp: localization offset; grasp: no grasp contact & 7 att.; 2 no change; rej.\ replay 2, regr 1, inv 1; infra/unrepl.\ 1 & routing refit & -- & branch numbering across nested calls; interpreter mounts in container\\
99 & F\,0/4 C\,0/4 S\,1/4 & find obj: target not found; find region: fixture not found; find region: wrong fixture; reach: tool path blocked & 8 att.; 3 no change; rej.\ fresh 3, replay 2 & routing refit & 2 not reg. & CUDA probe timeout; observer export detached\\
100 & C\,0/4 S\,0/4 & find obj: no task object near point; find obj: target not in view; grasp: closed above object; grasp: localization offset & 6 att.; 2 no change; rej.\ replay 3 & localize.\allowbreak{}sam3\_\allowbreak{}1\_\allowbreak{}da3 (find obj: target not in view); routing refit & 2 not reg. & nested-family branching; diffs skip tooling trees\\
101 & F\,0/4 C\,0/4 S\,1/4 & find region: wrong region; grasp: no grasp contact; move: tool path blocked; reach: tool path blocked & 6 att.; 3 no change; rej.\ replay 3 & routing refit & 1 not reg. & declared/commanded target mismatch is program fault\\
102 & F\,0/4 C\,0/4 S\,0/4 & find obj: no task object near point; find region: wrong fixture; grasp: no grasp contact & 8 att.; 2 no change; rej.\ replay 6 & routing refit & 1 not reg. & target mismatch triggers orchestrator revision; query naming a fixture is a program fault; partial rescues reach fresh-state gates; stall rotation follows fresh-state gates; proximal faults classify small gaps; orchestrator sees observed usage\\
103 & F\,0/4 C\,0/5 S\,0/4 & find region: wrong region; grasp: no grasp contact; move: tool path blocked; reach: tool path blocked & 10 att.; 1 no change; rej.\ replay 9 & routing refit & -- & no validation program: not applicable; requests under phi/requests recorded\\
104 & F\,0/4 C\,0/4 S\,0/4 & find region: wrong fixture; grasp: localization offset; move: tool path blocked; reach: tool path blocked & 9 att.; 2 no change; rej.\ replay 7 & routing refit & loftr\_\allowbreak{}multiview\_\allowbreak{}localization; 2 not reg. & identified mechanism ranks first; paired differences must reproduce\\
105 & F\,0/4 C\,0/4 S\,0/4 & find region: wrong region; grasp: no grasp contact; move: tool path blocked; reach: tool path blocked & 7 att.; 2 req.; 2 no change; rej.\ replay 3 & routing refit & 2 not reg. & measured rescue survives upstream evaluator change\\
106 & F\,0/4 C\,0/4 S\,0/4 & find obj: target not in view; find region: wrong fixture; grasp: no grasp contact; reach: tool path blocked & 8 att.; 1 req.; 1 no change; rej.\ regr 3, replay 2, fresh 1 & routing refit & equivariant\_\allowbreak{}grid\_\allowbreak{}base\_\allowbreak{}placement, relational\_\allowbreak{}rgbd\_\allowbreak{}localizer; 1 not reg. & probe suites import harness\\
107 & F\,0/4 C\,0/4 S\,1/4 & find obj: target not in view; find region: wrong region; move: tool path blocked; reach: tool path blocked & 7 att.; 2 req.; 1 no change; rej.\ replay 2, fresh 1 & localize.\allowbreak{}calibrated\_\allowbreak{}multiview\_\allowbreak{}rgb (find obj: target not in view); routing refit & observed\_\allowbreak{}serial\_\allowbreak{}pose\_\allowbreak{}transport; 2 not reg. & draw that cannot show repair does not reject\\
108 & F\,0/4 C\,0/4 S\,0/4 & find obj: target not in view; grasp: no grasp contact; move: tool path blocked; reach: tool path blocked & 8 att.; 2 req.; rej.\ fresh 4, replay 2 & routing refit & mplib\_\allowbreak{}whole\_\allowbreak{}body\_\allowbreak{}goalset; 2 not reg. & evaluator refs cached; task-span rule over exercised tasks\\
109 & F\,0/4 C\,0/4 S\,0/4 & find obj: target not in view; find region: wrong region; move: tool path blocked; reach: tool path blocked & 6 att.; 2 req.; 1 no change; rej.\ replay 3 & routing refit & observed\_\allowbreak{}serial\_\allowbreak{}pose\_\allowbreak{}transport; 1 not reg. & --\\
110 & F\,0/4 C\,0/4 S\,0/4 & find obj: target not in view; grasp: no grasp contact; move: tool path blocked; reach: tool path blocked & 8 att.; 2 req.; rej.\ replay 3, fresh 2, transfer 1 & routing refit & loftr\_\allowbreak{}multiview\_\allowbreak{}localization, mplib\_\allowbreak{}confined\_\allowbreak{}motion, observed\_\allowbreak{}serial\_\allowbreak{}pose\_\allowbreak{}transport, relational\_\allowbreak{}rgbd\_\allowbreak{}localizer; 1 not reg. & real wins count at transfer; requests never lost; truthful ledger; decline backoff; repair replayed for what was asked; provider hosts over ssh; grounding\_\allowbreak{}proposals capability class; memory budget above interpreter baseline; single-threaded program child\\
111 & F\,0/3 C\,0/4 S\,0/4 & find region: wrong fixture; grasp: no grasp contact; move: tool path blocked; reach: tool path blocked & 6 att.; 3 req.; rej.\ fresh 3 & routing refit & mplib\_\allowbreak{}whole\_\allowbreak{}body\_\allowbreak{}goalset; 7 not reg. & requests not judged by harness faults\\
112 & F\,0/4 C\,0/4 S\,0/4 & find region: wrong fixture; grasp: no grasp contact; move: tool path blocked; reach: tool path blocked & 6 att.; 3 req.; rej.\ fresh 2, transfer 1 & routing refit & equivariant\_\allowbreak{}grid\_\allowbreak{}base\_\allowbreak{}placement, observed\_\allowbreak{}serial\_\allowbreak{}pose\_\allowbreak{}transport; 4 not reg. & residual-matched sealed transfer states\\
113 & F\,0/4 C\,0/4 S\,0/4 & find obj: target not in view; grasp: no grasp contact; move: tool path blocked; reach: tool path blocked & 8 att.; 2 req.; rej.\ replay 3, fresh 3 & routing refit & loftr\_\allowbreak{}multiview\_\allowbreak{}localization, mplib\_\allowbreak{}confined\_\allowbreak{}motion, mplib\_\allowbreak{}free\_\allowbreak{}space\_\allowbreak{}reachability; 4 not reg. & --\\
114 & F\,0/4 C\,0/4 S\,0/4 & find obj: target not in view; find region: wrong region; move: tool path blocked; reach: tool path blocked & 6 att.; 3 req.; rej.\ replay 2, fresh 1 & routing refit & image\_\allowbreak{}expression\_\allowbreak{}semantic\_\allowbreak{}assessor, visual\_\allowbreak{}referent\_\allowbreak{}entailment; 4 not reg. & --\\
115 & F\,0/4 C\,0/4 S\,0/4 & find obj: target not in view; find region: wrong region; move: tool path blocked; reach: tool path blocked & 6 att.; 3 req.; rej.\ replay 3 & routing refit & mplib\_\allowbreak{}confined\_\allowbreak{}motion, observed\_\allowbreak{}transport\_\allowbreak{}assessment, visual\_\allowbreak{}referent\_\allowbreak{}entailment; 3 not reg. & --\\
116 & F\,0/4 C\,0/4 S\,0/4 & find obj: target not in view; find region: wrong region; move: tool path blocked; reach: tool path blocked & 6 att.; 3 req.; rej.\ transfer 1, replay 1, fresh 1 & routing refit & discriminative\_\allowbreak{}region\_\allowbreak{}association, discriminative\_\allowbreak{}region\_\allowbreak{}compatibility, mplib\_\allowbreak{}confined\_\allowbreak{}motion, relational\_\allowbreak{}rgbd\_\allowbreak{}localizer; 5 not reg. & --\\
117 & F\,0/4 C\,0/4 S\,0/4 & find region: wrong region; grasp: no grasp contact; move: tool path blocked; reach: tool path blocked & 6 att.; 2 req.; 1 no change; rej.\ fresh 3 & routing refit & mplib\_\allowbreak{}confined\_\allowbreak{}motion, observed\_\allowbreak{}transport\_\allowbreak{}assessment, relational\_\allowbreak{}rgbd\_\allowbreak{}localizer & --\\
118 & F\,0/4 C\,0/4 S\,0/4 & find region: wrong region; grasp: no grasp contact; move: tool path blocked; reach: tool path blocked & 4 att.; 3 req. & grasp.\allowbreak{}analytic\_\allowbreak{}top, grasp.\allowbreak{}rgbd\_\allowbreak{}6dof\_\allowbreak{}retain (grasp: no grasp contact); routing refit & mplib\_\allowbreak{}confined\_\allowbreak{}motion, observed\_\allowbreak{}region\_\allowbreak{}relational\_\allowbreak{}localizer, observed\_\allowbreak{}transport\_\allowbreak{}assessment, relational\_\allowbreak{}rgbd\_\allowbreak{}localizer; 1 not reg. & --\\
119 & F\,0/4 C\,0/4 S\,0/4 & find obj: target not in view; find region: wrong region; grasp: closed beside object; reach: target out of workspace & 10 att.; 1 req.; rej.\ replay 9 & routing refit & grounded\_\allowbreak{}clause\_\allowbreak{}program, observed\_\allowbreak{}region\_\allowbreak{}relational\_\allowbreak{}localizer, relational\_\allowbreak{}rgbd\_\allowbreak{}localizer, structured\_\allowbreak{}referent\_\allowbreak{}hypotheses; 3 not reg. & --\\
120 & F\,0/4 C\,0/4 S\,0/4 & find obj: target not in view; find region: wrong region; grasp: closed beside object; reach: tool path blocked & 8 att.; 3 req.; rej.\ replay 5 & routing refit & language\_\allowbreak{}spatial\_\allowbreak{}search\_\allowbreak{}memory, mplib\_\allowbreak{}confined\_\allowbreak{}motion, relational\_\allowbreak{}rgbd\_\allowbreak{}localizer; 4 not reg. & --\\
121 & F\,0/4 C\,0/4 S\,0/4 & find obj: target not in view; find region: wrong region; grasp: no grasp contact; reach: target out of workspace & 8 att.; 2 req.; rej.\ transfer 3, replay 2, fresh 1 & routing refit & discriminative\_\allowbreak{}observed\_\allowbreak{}pairs, selective\_\allowbreak{}context\_\allowbreak{}region\_\allowbreak{}assessor; 3 not reg. & --\\
122 & F\,0/4 C\,0/4 S\,0/4 & find obj: target not in view; find region: wrong region; grasp: closed above object; reach: target out of workspace & 10 att.; 1 req.; rej.\ replay 5, transfer 2, fresh 2 & routing refit & target\_\allowbreak{}graspgen\_\allowbreak{}6dof; 2 not reg. & --\\
123 & F\,0/4 C\,0/4 S\,0/4 & find obj: target not in view; find region: wrong region; grasp: no grasp contact; reach: tool path blocked & 10 att.; 1 req.; rej.\ replay 8, fresh 1 & routing refit & -- & --\\
\end{longtable}
\endgroup

\section{Measurements behind the main text}
\label{app:numbers}
The main text describes mechanisms; Table~\ref{tab:numbers} gives the measurements behind each statement. Table~\ref{tab:phases} summarizes how the set of tasks and the harness changed over the run. The first accepted change of the whole run was a new skill family for finding regions (round~10); the other accepted changes repaired or extended existing skills.

\begin{table}[h!]
\centering\small
\caption{\textbf{Measurements behind the main-text claims.} Rollout counts use completed rollouts; ``late'' means rounds 103--123; ``three-task phase'' means rounds 66--123.}
\label{tab:numbers}
\renewcommand{\arraystretch}{1.1}
\begin{tabularx}{\textwidth}{@{}p{.36\textwidth}X@{}}
\toprule
Claim & Measurement\\
\midrule
Target task never completed & \fridge\ 0/486; \cabinet\ 0/256; \sink\ 14/230, of which 9 had failed intermediate steps\\
Activity & 57 accepted library changes (51 perception); 86 provider registrations over 20 identities; 847 proposal attempts\\
Region grounding is the dominant first failure & First failed effect in 132/224 \fridge\ and 113/232 \cabinet\ rollouts (three-task phase); \code{wrong\_region} in all 62 late rollouts of fridge seeds 68, 70, 71\\
Semantic displacement & Region skill: 3,056 lines, 33 thresholds; composed localizer: 5,032 lines, 48 thresholds; 27 attempts on the two-state fridge gap rescued seed 69 in every trial and seed 68 in none\\
Visibility entanglement & Top-shelf occupant pixels at the region block's start: about 3,000 / 1,273 / 292 / 45 on seeds 68 / 71 / 70 / 69\\
VLM feasibility & Qwen2.5-VL 7B placed ``the top shelf'' and ``the shelf below'' correctly in 4/4 recorded head views, about 1\,s per query\\
Frontier lock & \fridge: 57\% fail step 1; 2.7\% pass four steps; 0.9\% pass five. In round 60, 5 of 6 fresh grasp cases never reached the grasp\\
Attenuated local win & Round-118 grasp change: reproduced wins on a fresh support state and a held-out state; next 60 rollouts: 0 task successes\\
Evaluator capture & 276 \code{reachable} successes at 0.849--0.850\,m under a 0.85\,m sphere; torso absent from the action description (34 proposer complaints); base speed advertised 0.0016\,m/step, measured 0.037\\
Unresolvable selection & 119 rejected candidates had rescued at least one exact state; 22 promotions (rounds 40--65) with 0 wins and 0 losses; 16 of 36 re-run disagreeing pairs did not reproduce\\
Evidence fixed point & Identical fridge and cabinet per-seed outcomes in rounds 90--102\\
Memory corruption & 88 request-bearing proposals gated as code (46 on the fridge-region front); gap declined 15 times citing the lost request\\
Capability without use & 11 provider identities first registered in rounds 114--121 have zero later collection calls\\
Framing sensitivity & 0/8 edits in rounds 93--94 vs.\ 11/12 in round 96 after one paragraph of proposer text changed\\
Human effort & 113 numbered harness fixes (Fig.~\ref{fig:prefix}b)\\
\bottomrule
\end{tabularx}
\end{table}

\begin{table}[h!]
\centering\small
\caption{\textbf{Phases of the run.} Boundaries summarize changes in which tasks were run and in the harness; they are not randomized treatment groups. ``Accepted'' counts skill changes that passed every test; routing refits are not included.}
\label{tab:phases}
\begin{tabularx}{\textwidth}{@{}lXrrrr@{}}
\toprule
Rounds & What changed & Rollouts & Completed & Task successes & Accepted changes\\
\midrule
1--59 & Fridge task only; discovery & 243 & 237 & 0 & 33\\
60--65 & Cabinet task added & 49 & 49 & 0 & 8\\
66--102 & Sink task added; harness repairs & 448 & 434 & 13 & 14\\
103--110 & Tests and memory repaired & 97 & 97 & 1 & 1\\
111--118 & Lost requests recovered; move to new machine & 96 & 95 & 0 & 1\\
119--123 & After the round-118 grasp change & 60 & 60 & 0 & 0\\
\midrule
Total & & 993 & 972 & 14 & 57\\
\bottomrule
\end{tabularx}
\end{table}

\section{External models and their use}
\label{app:providers}
Table~\ref{tab:provideruse} lists every external model the research agent registered. Models registered early (active viewing, base placement, grasp generation, motion planning, relational localization) were called by deployed skills thousands of times. Every model first registered from round~107 on has no call in any later task rollout. This includes all the vision-language grounders built for ``the top shelf'' (\S\ref{sec:perception}).

\begin{table}[h!]
\centering\small
\caption{\textbf{Every external model the research agent registered, and whether the robot used it.} Calls are counted in task rollouts of rounds after the model's first registration; test and research calls are not counted. Repairs of a model are folded into one row. The top block was used by deployed skills; the bottom block, mostly the late vision-language grounders for ``the top shelf'', was never called.}
\label{tab:provideruse}
\begin{tabular}{@{}lrrrr@{}}
\toprule
Model & First registered & Later rollouts & Calls & Rollouts with calls\\
\midrule
\path{target_free_active_view_base_placement} & 5 & 973 & 520 & 295\\
\path{equivariant_grid_base_placement} & 14 & 935 & 185 & 110\\
\path{target_graspgen_6dof} & 18 & 919 & 316 & 316\\
\path{relational_rgbd_localizer} & 19 & 915 & 1863 & 867\\
\path{mplib_whole_body_goalset} & 21 & 907 & 189 & 189\\
\path{mplib_confined_motion} & 23 & 899 & 51 & 45\\
\path{mplib_free_space_reachability} & 43 & 816 & 586 & 491\\
\path{loftr_multiview_localization} & 77 & 553 & 215 & 212\\
\path{observed_serial_pose_transport} & 107 & 192 & 0 & 0\\
\midrule
\path{visual_referent_entailment} & 114 & 108 & 0 & 0\\
\path{image_expression_semantic_assessor} & 114 & 108 & 0 & 0\\
\path{observed_transport_assessment} & 115 & 96 & 0 & 0\\
\path{discriminative_region_compatibility} & 116 & 84 & 0 & 0\\
\path{discriminative_region_association} & 116 & 84 & 0 & 0\\
\path{observed_region_relational_localizer} & 118 & 60 & 0 & 0\\
\path{structured_referent_hypotheses} & 119 & 48 & 0 & 0\\
\path{grounded_clause_program} & 119 & 48 & 0 & 0\\
\path{language_spatial_search_memory} & 120 & 36 & 0 & 0\\
\path{discriminative_observed_pairs} & 121 & 24 & 0 & 0\\
\path{selective_context_region_assessor} & 121 & 24 & 0 & 0\\
\bottomrule
\end{tabular}
\end{table}

\section{Where the run ended}
Table~\ref{tab:last} shows where each rollout of the final round stopped. Three fridge scenes fail at finding the top shelf before passing a single step; the fourth passes three steps and fails at the grasp. The cabinet task fails at finding its destination region, at grasping or at reaching. The sink task fails at seeing its target or at grasping.

\begin{table}[h!]
\centering\small
\caption{\textbf{Where each rollout stopped in the final round (123).} \emph{Steps passed} counts consecutive declared steps that succeeded before the first failure. No rollout succeeded.}
\label{tab:last}
\begin{tabular}{@{}llrll@{}}
\toprule
Task & Scene seed & Steps passed & First failed step & Diagnosis\\
\midrule
\fridge & 68 & 0 & \path{fixture_region_localized} & \path{wrong_region}\\
 & 69 & 3 & \path{retained} & \path{no_grasp_contact}\\
 & 70 & 0 & \path{fixture_region_localized} & \path{wrong_region}\\
 & 71 & 0 & \path{fixture_region_localized} & \path{wrong_region}\\
\cabinet & 68 & 3 & \path{retained} & \path{no_grasp_contact}\\
 & 69 & 1 & \path{fixture_region_localized} & \path{wrong_fixture}\\
 & 70 & 4 & \path{reachable} & \path{tool_path_blocked}\\
 & 71 & 1 & \path{fixture_region_localized} & \path{wrong_fixture}\\
\sink & 68 & 0 & \path{localized} & \path{target_not_in_view}\\
 & 69 & 2 & \path{retained} & \path{closed_above_object}\\
 & 70 & 2 & \path{retained} & \path{no_grasp_contact}\\
 & 71 & 2 & \path{retained} & \path{closed_above_object}\\
\bottomrule
\end{tabular}
\end{table}

\section{Additional figures}
\begin{figure}[h!]
\centering\includegraphics[width=\textwidth]{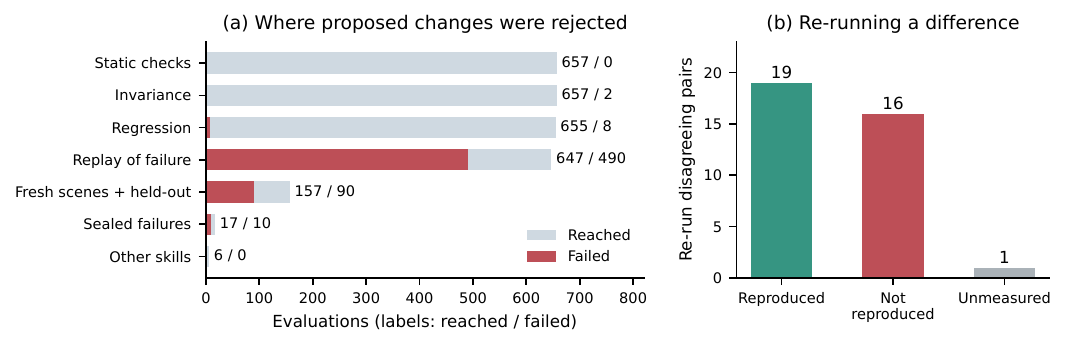}
\caption{\textbf{Where candidates were rejected, and how stable paired comparisons were.} (a) How many proposed changes reached and failed each test; replaying the recorded failure rejects most of them. (b) Of 36 fresh-scene pairs where the old and new skill disagreed and were re-run, 19 reproduced, 16 did not, and one was unmeasured.}
\label{fig:gates}
\end{figure}

\begin{figure}[h!]
\centering
\begin{minipage}{.31\textwidth}\centering
\includegraphics[width=.84\linewidth]{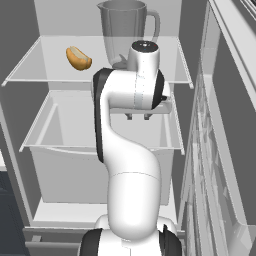}\\[-2pt]
\small (a) Fridge, seed 68, round 115
\end{minipage}\hfill
\begin{minipage}{.31\textwidth}\centering
\includegraphics[width=.84\linewidth]{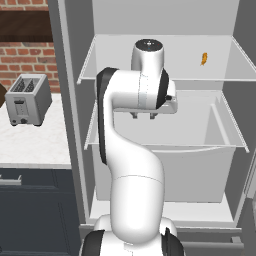}\\[-2pt]
\small (b) Fridge, seed 69, round 115
\end{minipage}\hfill
\begin{minipage}{.31\textwidth}\centering
\includegraphics[width=.84\linewidth]{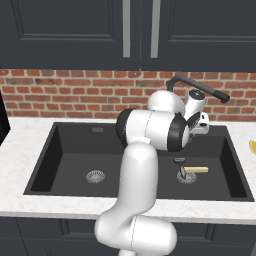}\\[-2pt]
\small (c) Sink, seed 71, round 107
\end{minipage}
\caption{\textbf{Recorded head-camera frames.} (a) Start of the top-shelf region block later judged \code{wrong\_region}: the lower shelf fills the view. (b) Start of a grasp block that fails with \code{no\_grasp\_contact}. (c) A late block of a native sink success whose program had an earlier \code{tool\_path\_blocked} failure. Frames are unmodified 256-pixel trace keyframes.}
\label{fig:frames}
\end{figure}

Figure~\ref{fig:gates} shows where proposed changes were rejected and how often paired differences between old and new skills reproduced when re-run. Most rejections come from replaying the exact recorded failure. Of the differences that were re-run, nearly half did not reproduce. Figure~\ref{fig:frames} shows recorded head-camera views from the fridge and sink tasks.

\end{document}